\documentclass{article}

\newcommand{\improve}[1]{{\color{gray}#1}}

\usepackage[dvipsnames]{xcolor}
\definecolor{myblue}{HTML}{00a0e1}
\definecolor{myorange}{HTML}{e6a532}

\newcommand{\mr}[1]{\mathit{#1}}

\newcommand{\mypar}[1]{\vspace{0pt}
\noindent\textbf{#1~}}

\usepackage{setspace}
\usepackage{makecell}
\usepackage{xcolor}
\usepackage{algorithm}
\usepackage{algorithmic}
\usepackage{glossaries}
\usepackage{stmaryrd}
\usepackage{multirow}
\usepackage{colortbl}
\usepackage{tikz}
\usepackage{etoc}
\usepackage{verbatim}
\usepackage{bm}

\makeglossaries
\newacronym{sota}{SOTA}{State Of The Art}
\newacronym{tta}{TTA}{Test-Time Adaptation}
\newacronym{ttt}{TTT}{Test-Time Training}
\newacronym{knn}{KNN}{K-Nearest Neighbors}
\newacronym{fps}{FPS}{Farthest Point Sampling}
\newacronym{ss}{§§}{Supplementary Section}
\newacronym{pg}{PG}{Purge-Gate}
\newacronym{pg-sp}{PG-SP}{Source-Prototype based Purge-Gate}
\newacronym{pg-sf}{PG-SF}{Source-Free Purge-Gate}

\newcommand{\mybf}[1]{#1}

\newcommand{\mybs}[1]{#1}

\newcommand{\xx}{\mybf{x}}
\newcommand{\XX}{\mybf{X}}
\newcommand{\ddelta}{\mybf{\delta}}
\newcommand{\cc}{\mybf{c}}
\newcommand{\pp}{\mybf{p}}
\newcommand{\uu}{\mybf{u}}
\newcommand{\vv}{\mybf{v}}

\newcommand{\mmu}{\mybs{\mu}}
\newcommand{\ssigma}{\mybs{\sigma}}
\newcommand{\SSigma}{\mybs{\Sigma}}

\newcommand{\NN}{\mathcal{N}}
\newcommand{\EE}{\mathcal{E}}

\usepackage{color, colortbl}

\usepackage[dvipsnames]{xcolor}
\colorlet{MyCol}{Tan!10}

\newcommand{\colrow}{\rowcolor{MyCol}}

\newcommand{\phz}{\phantom{0}}

\newcommand{\scdot}{\!\cdot\!}

\newcommand{\tr}[1]{{#1}^{\top}}

\usepackage{amsmath}
\usepackage{amssymb}
\usepackage{graphicx,scalerel}

\newcommand{\CLS}{[\texttt{cls}]}

\usepackage{arxiv}

\usepackage[utf8]{inputenc} 
\usepackage[T1]{fontenc}    
\usepackage{hyperref}       
\usepackage{url}            
\usepackage{booktabs}       
\usepackage{amsfonts}       
\usepackage{nicefrac}       
\usepackage{microtype}      
\usepackage{cleveref}       
\usepackage{lipsum}         
\usepackage{graphicx}
\usepackage{natbib}
\usepackage{doi}
\usepackage{subcaption}

\title{Purge-Gate: Backpropagation-Free Test-Time Adaptation for Point Clouds Classification via Token Purging}

\newif\ifuniqueAffiliation
\uniqueAffiliationtrue

\ifuniqueAffiliation 
\author{%
Moslem Yazdanpanah\thanks{Correspondence to \href{mailto:moslem.yazdanpanah.1@ens.etsmtl.ca}{moslem.yazdanpanah.1@ens.etsmtl.ca}} \\
LIVIA, ÉTS Montréal, Canada \\
International Laboratory on Learning Systems (ILLS) \\
\texttt{moslem.yazdanpanah.1@ens.etsmtl.ca} \\
\And
Ali Bahri \\
LIVIA, ÉTS Montréal, Canada \\
International Laboratory on Learning Systems (ILLS) \\
\And
Mehrdad Noori \\
LIVIA, ÉTS Montréal, Canada \\
International Laboratory on Learning Systems (ILLS) \\
\And
Sahar Dastani \\
LIVIA, ÉTS Montréal, Canada \\
International Laboratory on Learning Systems (ILLS) \\
\And
Gustavo Adolfo Vargas Hakim \\
LIVIA, ÉTS Montréal, Canada \\
International Laboratory on Learning Systems (ILLS) \\
\And
David Osowiechi \\
LIVIA, ÉTS Montréal, Canada \\
International Laboratory on Learning Systems (ILLS) \\
\And
Ismail Ben Ayed \\
LIVIA, ÉTS Montréal, Canada \\
International Laboratory on Learning Systems (ILLS) \\
\And
Christian Desrosiers \\
LIVIA, ÉTS Montréal, Canada \\
International Laboratory on Learning Systems (ILLS) \\
}
\else
\usepackage{authblk}

\fi

\renewcommand{\undertitle}{Accepted at ICCV 2025}

\hypersetup{
  pdftitle={Purge-Gate: Backpropagation-Free Test-Time Adaptation for Point Clouds via Token Purging},
  pdfsubject={cs.CV, cs.LG},
  pdfauthor={Moslem Yazdanpanah, Ali Bahri, Mehrdad Noori, Sahar Dastani, Gustavo Adolfo Vargas Hakim, David Osowiechi, Ismail Ben Ayed, Christian Desrosiers},
  pdfkeywords={test-time adaptation, point clouds, domain shift, backpropagation-free methods, token purging, 3D vision, robustness, transformers, deep learning}
}

\begin{document}
\maketitle

\begin{abstract}
	Test-time adaptation (TTA) is crucial for mitigating performance degradation caused by distribution shifts in 3D point cloud classification. In this work, we introduce Token Purging (PG), a novel backpropagation-free approach that removes tokens highly affected by domain shifts before they reach attention layers. Unlike existing TTA methods, PG operates at the token level, ensuring robust adaptation without iterative updates. We propose two variants: PG-SP, which leverages source statistics, and PG-SF, a fully source-free version relying on CLS-token-driven adaptation. Extensive evaluations on ModelNet40-C, ShapeNet-C, and ScanObjectNN-C demonstrate that PG-SP achieves an average of +10.3\% higher accuracy than state-of-the-art backpropagation-free methods, while PG-SF sets new benchmarks for source-free adaptation. Moreover, PG is 12.4× faster and 5.5× more memory efficient than our baseline, making it suitable for real-world deployment. Code is available at \hyperlink{https://github.com/MosyMosy/Purge-Gate}{https://github.com/MosyMosy/Purge-Gate}
\end{abstract}

\keywords{Point Cloud \and Test Time Adaptation \and More}

\section{Introduction}
\label{sec:introduction}

Deep neural networks have recently achieved remarkable success in classifying 3D point clouds \cite{qi2017pointnet, qi2017pointnet++, wang2019dynamic, pang2022masked, zhang2022point, zhang2023learning, bahri2024geomask3d}. However, their performance can degrade significantly when the test distribution (\emph{target}) differs from the training distribution (\emph{source}). In 3D data, such shifts can arise from various factors, including differences in sensor types and occlusions. Given the vast range of possible distribution shifts, pretraining a model for every scenario is impractical. 
\begin{figure}
    \centering
    \includegraphics[width=1\linewidth]{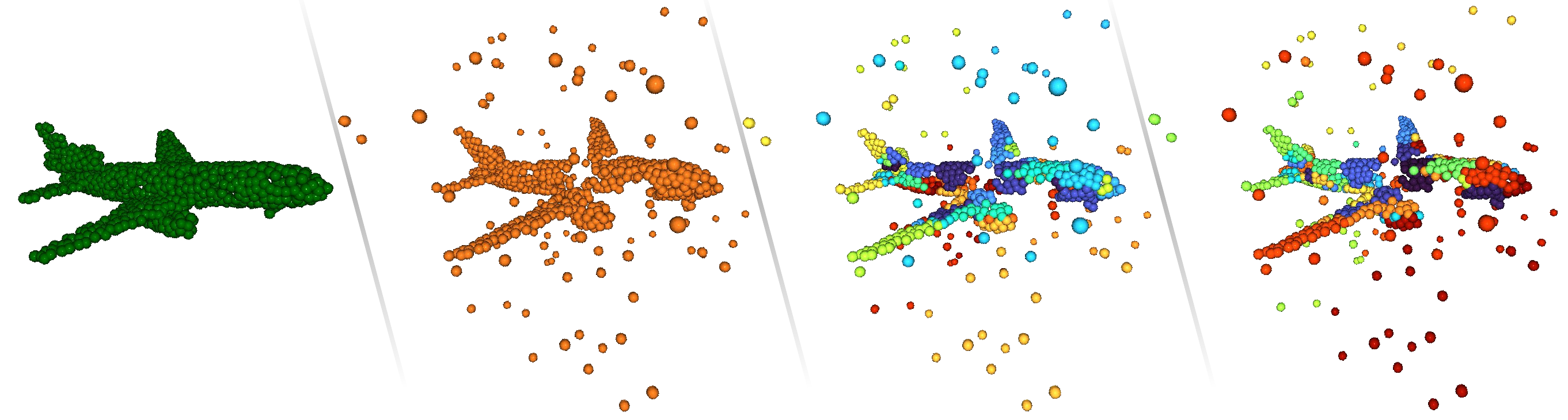}
    \caption{Token-level domain shift visualization. From left to right: clean point cloud, corrupted point cloud, tokens ranked by \gls{pg-sf}, and tokens ranked by \gls{pg-sp}. The color gradient from red to green indicates the degree of domain shift, with red representing the highest shift and green the lowest.}
    \label{fig:plane}
    \vspace{-10pt}
\end{figure}
\acrfull{tta} mitigates distribution shifts that arise after training \cite{gandelsman2022test, gao2023back, iwasawa2021test, lim2023ttn, yeo2023rapid, schneider2020improving} by enabling a source-pretrained model to adapt dynamically during inference using only unlabeled test data. Along with light weight transfer learning methods \cite{yazdanpanah2022revisiting, yazdanpanah2022visual}, various \gls{tta} methods have been proposed for 2D image analysis \cite{osowiechi2024nc, osowiechi2024watt, hakim2025clipartt, noori2025test}, including techniques that adjust batch statistics and affine parameters through entropy minimization \cite{nado2020evaluating,wang2020tent}, optimize a Laplacian-based clustering loss \cite{boudiaf2022parameter}, or average the output distributions of multiple augmentations of individual test samples \cite{zhang2022memo}. In contrast to \gls{tta} in the image modality, where we assume a fixed-dimensional input with a uniformly distributed target noise over the image area, test-time noise in point clouds can manifest in more complex ways. It may introduce new points, thereby expanding the dimensionality of the point cloud, or appear as unevenly distributed perturbations that affect the original structure. Consequently, applying similar adaptation strategies to point clouds presents unique challenges due to their non-Euclidean nature and inherent structural sparsity.

In recent years, numerous studies have focused on developing \gls{tta} methods for point cloud classification \cite{wang2024backpropagation,mirza2023mate,wang2024continual,bahri2024test}. While these approaches have shown promising results, they come with notable limitations. For example, TENT \cite{wang2020tent}, as a dominant approach for image \gls{tta}, performs poorly when applied to point clouds. MATE \cite{mirza2023mate} employs a Masked Autoencoder (MAE) reconstruction objective for both model pretraining and test-time adaptation. As a result, it falls under \gls{ttt} approaches, which depend on auxiliary training-time tasks and cannot be applied to arbitrarily pre-trained models. Similarly, the method in \cite{bahri2024test} adapts the model using multiple-point sub-sampling augmentations, significantly increasing computational cost. Finally, these methods iteratively adapt the model through multiple backpropagation steps. Consequently, they are unsuitable for runtime-critical applications and are sensitive to the number of update steps. BFTT3D \cite{wang2024backpropagation} was recently proposed as a backpropagation-free method for \gls{tta} of point cloud data. Despite its adaptation-free advantages, it relies heavily on detailed class-level source prototypes, making it dependent on source data during the test time.

On the other hand, a recent line of research has focused on improving the efficiency of transformer \cite{vaswani2017attention} networks by optimizing the number of tokens, a technique known as Token Pruning. Introduced by \cite{kim2022learned}, these methods aim to identify and remove redundant or uninformative tokens. By eliminating these unnecessary tokens, they enhance the inference efficiency of transformers without altering the backbone parameters.
Inspired by Token Pruning and motivated by the challenge posed by domain shift, we introduce an innovative application of token filtering to the task of \gls{tta}. Our approach focuses on detecting and removing tokens that are highly affected by domain shift at test time. Unlike Token Pruning, which targets unimportant tokens to improve computational efficiency, our method identifies tokens most impacted by domain shift 
\cref{fig:plane} 
and enhances \gls{tta} performance. \textit{To emphasize its focus on filtering corrupted tokens, we refer to our approach as Token Purging}. 
Unlike existing methods, which have the above-mentioned limitations, our approach mitigates distribution shift at the token level. We observe that noise from domain shifts disrupts the attention mechanism in transformer-based architectures, impairing feature aggregation. To counter this, we propose \gls{pg}, a lightweight, backpropagation-free mechanism that detects 
and filters noisy tokens before they reach the attention layers. By dynamically refining the token set at inference time, our method improves performance under domain shifts without requiring additional optimization steps.

\noindent Our contributions can be summarized as follows:
\begin{itemize}\setlength\itemsep{.25em}
    \item We formalize the effect of domain shift on the attention mechanism in point cloud Transformers. Based on this, we introduce \gls{pg}, a backpropagation-free mechanism that selectively removes highly divergent tokens, thus improving test-time adaptation performance. 
    \item We propose two variants of \gls{pg}—\gls{pg-sp}, which utilizes source statistics when available, and \gls{pg-sf}, operating in a fully source-free manner. 
    \item We validate our approach across multiple benchmarks, including datasets with standard synthetic corruption and real-world datasets with domain shift (for the first time), using Transformer \cite{vaswani2017attention} and Mamba \cite{gu2023mamba} (for the first time) architectures. We show that our method outperforms existing \gls{tta} methods while maintaining computational efficiency.
\end{itemize}

By addressing the challenges specific to point cloud adaptation, our method advances the field of test-time adaptation and offers a practical solution for improving model robustness in 3D \acrfull{tta}.


\section{Related Works}
\label{sec:related}

\mypar{Test-Time Point Cloud Adaptation.}
Research on \acrfull{tta} for 3D point clouds remains limited compared to 2D image domains. One of the earliest methods in this area, MATE \cite{mirza2023mate}, employs a Masked Autoencoder (MAE) pretraining objective to enhance model robustness against distribution shifts. Following MATE, SMART-PC \cite{Bahri2025SMARTPC}, has proposed learning the object's skeleton as an auxiliary task to update the network during the training time. However, since MATE and SMART-PC rely on auxiliary training-time tasks, they fall under \acrfull{ttt} rather than pure \gls{tta}. Similarly, MM-TTA \cite{shin2022mm} integrates predictions from multiple modalities to reinforce self-supervised signals in 3D semantic segmentation. While effective in multimodal settings, its reliance on multiple input modalities makes it less applicable to single-modality point cloud classification.

BFTT3D \cite{wang2024backpropagation} introduces a backpropagation-free \gls{tta} method specifically designed for 3D data. It mitigates domain shifts through a two-stream architecture that retains both source and target domain knowledge. However, it relies on class-specific prototypes derived from the source data, making it incompatible with fully source-free \gls{tta} settings.

Unlike these methods, our approach does not require auxiliary training objectives, multimodal data, or reliance on class-level source prototypes. Instead, we introduce a token-level adaptation mechanism that directly mitigates the effects of domain shift during inference in a backpropagation-free manner.

\mypar{Token Pruning.}
Token pruning is a recent technique designed to improve the efficiency of transformer models by reducing the number of tokens during inference \cite{kim2022learned}. DToP \cite{tang2023dynamic} enhances semantic segmentation efficiency by dynamically pruning tokens based on their difficulty, enabling early exits while preserving contextual information. The Pruning \& Pooling Transformers (PPT) framework \cite{wu2023ppt} optimizes Vision Transformers (ViTs) by combining token pruning and pooling, thereby reducing computational cost without requiring additional trainable parameters.
While these approaches primarily aim to improve inference speed by eliminating redundant tokens, they do not address the challenge of domain shifts. Our method takes a different direction: rather than pruning unimportant tokens for efficiency, we identify and remove tokens most affected by distribution shift, thereby enhancing \gls{tta} performance in a backpropagation-free manner.

\mypar{Mamba Architecture.} Mamba \cite{gu2023mamba} is a variation of State Space Models that is introduced to the computer vision field by the VMamba paper \cite{liu2024vmamba}. Later, PointMamba \cite{liang2024pointmamba} adapted this architecture to point cloud modality. There are numerous works aimed at enhancing this architecture \cite{dastani2025spectral, bahri2025spectral}, but no empirical evaluation in a standard \gls{tta} setting. To show the versatility of our method, for the first time, we have used the Mamba network in a point cloud \gls{tta} setting.


\section{Method}
\label{sec:method}
\begin{figure*}
    \centering
    \includegraphics[width=1\linewidth]{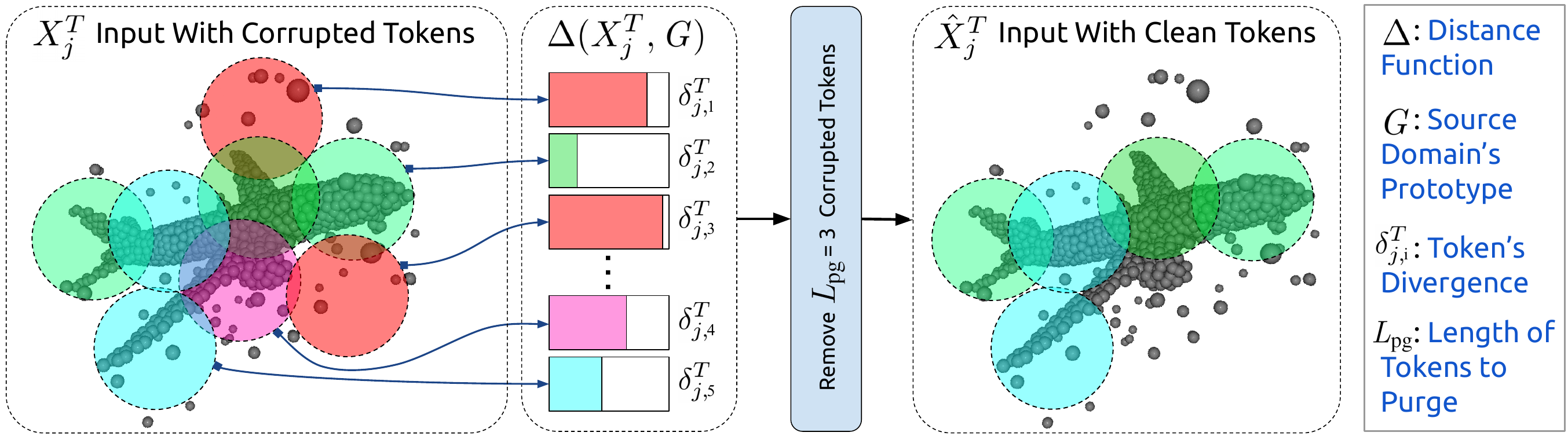}
    \caption{The overall procedure of \gls{pg} algorithm at the input of the attention layer.
    }
    \label{fig:main_fig}
    \vspace{-10pt}
\end{figure*}
A point cloud is an unordered set of 3-D points
    $\XX = \{\pp_{1}, \pp_{2}, \dots, \pp_{N_p}\} \subset \mathbb{R}^{3}$,
where \(N_p\) is the number of points.
In the point cloud modality, we follow \cite{pang2022masked} to tokenize the point cloud samples 
$\XX_i = (\xx_{i,1}, \xx_{i,2}, \dots, \xx_{i,\mr{L}_t})$. 
We form each input token $\xx_{i,t} = (\cc_t, \NN_t)$ by selecting a center point $\cc_t$ and its neighborhood $\NN_t$ as follows:
\begin{align}
  \cc_t \, &= \, \arg \max_{\pp \in \XX} \, \min_{\cc' \in \{\cc_1, \dots, \cc_{t-1}\}} \, \| \pp - \cc' \|_2, \\
  \NN_t \, &= \, \bigl\{ \pp \in \XX \,\big|\, \pp \in \mr{KNN}(\cc_t) \bigr\},
\end{align}
Here, $t$ denotes the token index and $\mr{L}_t$ is the total number of tokens for each point cloud. The center point $\cc_t$ is chosen via a furthest-point strategy, and $\mr{KNN}(\cc_t)$ selects the $k$-nearest neighbors of $\cc_t$. This tokenization preserves the local neighborhood structure in the 3D point cloud, which is beneficial for subsequent attention-based layers.

Let the source domain consist of labeled training data:
\begin{equation}
    \mathcal{D}_S = \bigl\{\bigl(\XX_i^S, y_i\bigr)\bigr\}_{i=1}^{N_S},
\end{equation}
where $y_i$ is the label of the $i$-th source sample, and $N_S$ is the total number of source training samples. The model $f_{\theta}$ has three main components: 
a mapping $\EE$ 
transforming each token $\xx_{i,t}$ into an embedded feature vector $\EE(\xx_{i,t}) \in \mathbb{R}^d$; a stack of transformer-based layers that model interactions among the embedded tokens; and an output head that produces classification logits $\hat{y}_i = f_{\theta}(\XX_i)$ based on the final layer’s representations.
For brevity, \textit{we continue to use $\XX$, and $\xx$ to denote the intermediate embedded sample and token throughout the network layers}.

At test time, the model encounters unlabeled target domain data:
\begin{equation}
    \mathcal{D}_T = \bigl\{\XX_j^T \bigr\}_{j=1}^{N_T}, \\
      \quad 
      \XX_j^T = \big(\xx_{j,1}^T, \xx_{j,2}^T, \dots, \xx_{j,\mr{L}_t}^T\big),
\end{equation}
whose distribution differs from the source domain, i.e., $P_S(\XX) \neq P_T(\XX)$. In this work, we assume that this domain shift can be modeled as an unknown noise function ${\epsilon}$ in the embedding space.

During test-time adaptation, our goal is to find an optimal $f_{\theta'}$ that maximizes classification accuracy on $P_T$.

A straightforward solution to remedy the performance drop under distribution shift is to fine-tune $\theta$ on $\mathcal{D}_T$ using a label-free objective such as entropy minimization. However, this approach becomes suboptimal when the target domain differs significantly from the source, as the model's incorrect predictions are reinforced, ultimately leading to its collapse. Instead, we adopt a different approach centered on the effect of noise on the attention mechanism in transformer architectures.

\subsection{Attention Under Distribution Shift}

In each self-attention layer, the interaction between model weights $\theta$ and token embeddings $\xx_{i,t}$ determines how information is aggregated. In this section, we examine how this interaction can be perturbed by test-time noise, and then use this analysis to derive our \gls{pg} mechanism. 

Let $x_{j,t}^T = x_{j,t}^S + \epsilon_{j,t}$ be a noisy token where $x_{j,t}^S$ is the clean embedding and $\epsilon_{j,t}$ is an unknown perturbation such that $\|\epsilon_{j,t}\| = \|x_{j,t}^S - x_{j,t}^T\|$. Transformer layers apply per-token layer normalization (LN), ensuring zero mean and unit variance of normalized features:
\begin{equation}
\label{eq:ln}
    \mr{LN}\big(x_{j,t}^T\big)
  \;=\;
  \gamma \,
  \frac{\,x_{j,t}^T \;-\; \mu\bigl(x_{j,t}^T\bigr)}{\sigma\bigl(x_{j,t}^T\bigr)}
  \;+\;
  \beta
\end{equation}
where $\gamma,\beta$ are trainable parameters. We build on the following propositions to show how noisy tokens degrade the attention mechanism of ViTs.

\mypar{Proposition (Local Lipschitz Continuity of LN).}\emph{%
Let $x\!\in\!\mathbb{R}^d$ and define $\mr{LN}$ as in \cref{eq:ln}.  
Assume we restrict attention to a compact set 
$\mathcal{D}\subset\mathbb{R}^d$ on which the per‑token
standard deviation satisfies $\sigma(x)\neq0$ (equivalently,
$\sigma(x)\ge\sigma_{\min}>0$ for all $x\in\mathcal{D}$).
Then $\mr{LN}$ is \emph{locally Lipschitz} on $\mathcal{D}$: there exists a
finite constant $M>0$ such that for any $x_{j,t}^S, x_{j,t}^T\in\mathcal{D}$,
\begin{equation}
\label{eq:lipschitz}
    \bigl\|\mr{LN}(x_{j,t}^S)-\mr{LN}(x_{j,t}^T)\bigr\|
    \;\le\;
    M\,\|\epsilon_{j,t}\|.
\end{equation}}%

\mypar{Sketch of Proof.}\emph{%
Away from $\sigma(x)=0$, $\mr{LN}$ is a composition of smooth
maps:
\textbf{(1)} $x\mapsto\mu(x)$ is linear;
\textbf{(2)} $x\mapsto\sigma(x)$ is smooth where $\sigma(x)>0$;
\textbf{(3)} $x\mapsto1/\sigma(x)$ is smooth on $(0,\infty)$;
\textbf{(4)} the affine map $x\mapsto\gamma x+\beta$ is smooth.
Hence $\mr{LN}$ is continuously differentiable on any subregion that
stays clear of $\sigma(x)=0$.  
Because $\mathcal{D}$ is compact and contained in that region,
the Jacobian $J_{\mr{LN}}(x)$ is continuous and attains a finite
maximum, say $M=\max_{x\in\mathcal{D}}\|J_{\mr{LN}}(x)\|$.
Applying the mean‑value theorem to any $x_{j,t}^S, x_{j,t}^T \in\mathcal{D}$ yields
\[
   \|\mr{LN}(x_{j,t}^S)-\mr{LN}( x_{j,t}^T)\|
   \;\le\;
   \|J_{\mr{LN}}(\xi)\|\,\|\epsilon_{j,t}\|
   \;\le\;
   M\,\|\epsilon_{j,t}\|,
\]
for $\xi$ on the line segment between $u$ and $v$,
confirming \eqref{eq:lipschitz}.}\vspace{2pt}

For $x \in\mathcal{D}$, the noisy token $ x_{j,t}^T$ with small perturbations stays close to the clean
one due to the normalization. But when noise dominates, i.e. $\|\epsilon_{j,t}\|\gg\|x_{j,t}^S\|$, token values approximate random variables leading to $\mr{LN}(x_{j,t}^T) \approx \mr{LN}(\epsilon_{j,t})$.

\mypar{Proposition (Random Directions in High Dimension).}
\emph{Let $\uu,\vv\!\in\!\mathbb{R}^{d^{'}}$ be random vectors drawn from a sufficiently isotropic distribution on a sphere $\mathbb{S}^{d'-1}$. Then for large $d'$,
\begin{equation}
\label{eq:sphere}
    \mathbb{E}\bigl[\uu^\top \vv\bigr] \;\approx\; 0,
  \ \ \text{and} \ \
  \mathrm{Var}\bigl(\uu^\top \vv\bigr) \;\approx\; \frac{1}{d'}.
\end{equation}
}

\mypar{Sketch of Proof.}\emph{
When $\uu,\vv$ are i.i.d.\ random points on $\mathbb{S}^{d-1}$, the distribution 
of their dot product is sharply concentrated near zero, due to classical concentration-of-measure 
results. Intuitively, distinct random directions in high dimensions tend to be nearly orthogonal. 
Hence, the mean and variance of $\uu^\top \vv$ scale as above, with the variance 
shrinking on the order of $1/d'$.
}

Similarly, as $\mr{LN}$ enforces zero mean and unit variance per token, random (noise dominant) high-dimensional embeddings after $\mr{LN}$ effectively lie near a sphere of $\sqrt{d}$ (as $\mr{LN}$ ensures each vector \(x_{j,t}^T\) has zero mean and unit variance \textit{across its coordinates}, giving \(\|x_{j,t}^T\|\approx\sqrt{d}\)). 

Hence, from \cref{eq:sphere}, such noise-dominant tokens $x_{j,t'}^T$ move the attention score to a very small value
\begin{equation}
    S_{ij} = \tr{\big(Q_{j,t}^T\big)\!}K_{j,t'}^T \,\approx 0, \quad t' \neq t
\end{equation}
where $Q_{j,t}^T \!=\! \mr{LN}(x_{j,t}^T)  W_Q$,  $K_{j,t'}^T \!=\! \mr{LN}(x_{j,t'}^T)  W_K$, while $ W_Q,  W_K \in \mathbb{R}^{d\times d}$ and $Q_{j,t}^T, K_{j,t'}^T \in \mathbb{R}^d$.

As noise magnitude increases, attention scores $S_{ij}$ become closer to zero. In this situation, the softmax function yields near-identical attention weights in the attention layer, 
\begin{equation}
\label{eq:uniform}
    A_{ij} = \frac{\exp(S_{ij} / \sqrt{d})}{\sum_k \exp(S_{ik} / \sqrt{d})}
    ~\approx~
\frac{1}{\mr{L}_t},
\end{equation}
leading to ineffective uniform attention. From \cref{eq:lipschitz} and \cref{eq:uniform}, the higher the domain shift, the more it affects the ViT's attention mechanism.  

\subsection{\acrfull{pg}}
Based on this understanding, we propose \acrfull{pg}, a backpropagation-free gating mechanism to filter out the highly affected tokens at the input of the attention layer. As illustrated in \cref{fig:main_fig}, having a set of target tokens $\XX_j^T$, our PG mechanism calculates the token's divergence using a distance function $\Delta$ between the target tokens and a source domain's prototype $\mr{G}$:
\begin{equation}
    \Delta(\XX_j^T, \mr{G}) = (\ddelta_{j,1}^T, \ddelta_{j,2}^T, \dots, \ddelta_{j,\mr{L}_t}^T).
\end{equation}
Then, based on the token level distances $\ddelta_{j,t}^T$, it removes the top divergent tokens from $\XX_j^T$ forming a purified input $\hat{\XX}_j^T$ which contained a subset of the sample's tokens
\begin{equation}
    \hat{\XX}_j^T = \xx_j^T \setminus 
    \big\{ x_{j,t}^T \mid t \in \arg \max_{\mr{S} \subseteq [\mr{L}_t], |\mr{S}| = \mr{L}_{\text{pg}}} 
    \sum_{t \in \mr{S}} \ddelta_{j,t}^T \big\}
\end{equation}
where $\mr{L}_{\text{pg}}$ is the purging size and $\mr{\hat{L}}_{t} = \mr{L}_{t} - \mr{L}_{\text{pg}}$. Having a fixed-size $\mr{L}_{\text{pg}}$, is essential to preserve the Batch-Wise parallel computation.

Although this purging procedure seems straightforward, due to the unknown nature of the noise and its distribution, it is still challenging to find a good prototype $\mr{G}$. From this, we propose two variants of Purge-Gate: A \acrfull{pg-sp} and a \acrfull{pg-sf}.

\subsubsection{\acrfull{pg-sp}}
From \cref{eq:lipschitz}, we observe that $\mathcal{D}_S$ serves as a suitable basis for the prototype $\mr{G}$. We can acquire this prototype from the source data in two different ways:

\mypar{\emph{a})} By extracting the mean and variance statistics of the source data embeddings. This can be achieved in a computationally and memory-efficient manner using Welford's online algorithm, as follows:
\begin{align}\label{eq:weldorf}
    \mmu_i & \, = \, \frac{1}{L_t} \sum_{t=1}^{L_t} \EE\big(\xx_{i,t}^S\big),\\
    \ssigma_i & \, = \, \sqrt{\frac{1}{L_t} \sum_{t=1}^{L_t} \left(\EE\big(\xx_{i,t}^S\big) - \mmu_S\right)^2},\\
    \phi_S  & \leftarrow \phi_S + \frac{1}{n}\bigl(\phi_i-\phi_S\bigr),\qquad
                \phi \in \{\mmu,\ssigma\}. \label{eq:weldorf-mean}
\end{align}

\mypar{\emph{b})} By accumulating these statistics in the first LayerNorm (LN), formulated in  \cref{eq:ln}, in a source-free mode. Although the LN does not store these statistics by default, with a slight modification, it can accumulate and store such training-time source statistics for test-time inference. In this modified LN layer, we calculate such statistics using the Welford online algorithm of \cref{eq:weldorf-mean} at the token level.

In both \textbf{\emph{a})} and \textbf{\emph{b})} cases, $\mr{G} = (\mmu_S, \ssigma_S)$ and $\mmu_S, \ssigma_S \in \mathbb{R}^d$. Based on this definition of $\mr{G}$, we define $\Delta$ as the Mahalanobis distance between target tokens $x_{j,t}^T$ and $\mr{G}$:
\begin{equation}
\Delta(\XX_j^T, \mr{G}) = \sqrt{(\XX_j^T - \mmu_S)^T \SSigma_S^{-1} (\XX_j^T - \mmu_S)},
\end{equation}
where $ \SSigma_S = \text{diag}(\ssigma_S^2) $ is the diagonal covariance matrix derived from the stored variance values.

\subsubsection{\acrfull{pg-sf}}
Although acquiring the prototype $\mr{G}$ directly from source domain samples provides an accurate way of measuring the target domain shift, this approach still needs access to the source data (\acrshort{pg-sp}-\textbf{\emph{a}}) or retraining the model with modified LN layers (\acrshort{pg-sp}-\textbf{\emph{b}}). In real-world scenarios, such constraints are restrictive. To address this, we focus on a different approach to find a prototype $\mr{G}$. 

Each classification transformer $f_{\theta}$ contains a learnable $\CLS \in \mathbb{R}^d$ token, which is updated in pretraining such that its transformed representation effectively \emph{summarizes} the content of the other tokens in each training example. In the Transformer layer, the self-attention mechanism updates $ \CLS $ based on its interaction with the other tokens. Given an input sequence represented as $ \XX_i^S = (\CLS, \xx_{i,1}^S, \dots, \xx_{i,L_t}^S) \in \mathbb{R}^{(L_t+1) \times d} $, the corresponding query, key, and value matrices are computed as $Q_i = \mr{LN}(\XX_i^S) W_Q$, $K_i = \mr{LN}(\XX_i^S) W_K$, and $V_i = \mr{LN}(\XX_i^S) W_V$.
The $\CLS$ token update is determined by the self-attention mechanism, where the new representation of $ \CLS $ is computed as 
\begin{equation}
    \CLS' = \sum_{t=1}^{L_t} \alpha_{0t} V_{i,t},
\end{equation}
with attention weights defined as 
\begin{equation}
    \alpha_{0t} = \frac{\exp(S_{0t} / \sqrt{d})}{\sum_{t'=1}^{L_t} \exp(S_{0t'} / \sqrt{d})}, 
    \ \
    S_{0t} = \tr{Q_{i,0}\!}K_{i,t}.
\end{equation}
The gradients computed in the pre-training objective force the randomly-initialized $ \CLS $ to attend to relevant input features by adjusting attention representations and refining self-attended aggregation. Over time, this optimization process causes $ \CLS $ to converge toward a learned prototype of the source token distribution, effectively summarizing the key properties of the training-time embeddings. This aligns with previous studies \cite{zou2024closer2} reporting that “the CLS token naturally absorbs domain information”.

Based on this, in \acrshort{pg-sf}, we set the prototype $\mr{G}$ during test time as:
\begin{equation}
    \mr{G} = \mr{LN}\big(\CLS\big) W_Q, \ \ \mr{G} \in \mathbb{R}^d
\end{equation}
Since $\mr{G}$ is a single element prototype projected to the Query space, we use a cosine distance to calculate the shift between this prototype and projected target tokens
\begin{equation}
\Delta(\mathbf{X}_j^T, \mr{G}) = - \frac{
{\big(\mr{LN}\big(\mr{\mathbf{X}_j^T}\big) W_K \big)}\mr{G}
}
{\| \left(\mr{LN}\big(\mr{\mathbf{X}_j^T}\big) W_K \right)\| \scdot \|\mr{G}\|}.
\end{equation}

\subsection{Backpropagation-Free Considerations}
\subsubsection{Test-Time Hyperparameter Selection} In the \gls{pg} algorithm and its variants, \gls{pg-sp} and \gls{pg-sf}, the purging length hyperparameter $\mr{L}_{\text{pg}}$ plays a crucial role. Since these algorithms operate without backpropagation, $\mr{L}_{\text{pg}}$ cannot be learned directly. Moreover, its optimal value is noise-dependent, varying across different corruptions, making exhaustive selection infeasible during \gls{tta}. 

To address this, we employ a parallel test-time hyperparameter selection strategy. Specifically, during test-time evaluations, we execute the \gls{pg} algorithm for a predefined subset of $\mr{L}_{\text{pg}}$ values, denoted as $\mathcal{L}_{\text{pg}} = \{L_1, L_2, \dots, L_K\}$, and select the logit with the minimum entropy:
\begin{equation}
    L_{\text{pg}}^* = \arg\min_{L \in \mathcal{L}_{\text{pg}}} H(f(x; L))
\end{equation}
where $ f(x; L) $ denotes the output logits of the model given input $ x $ and purging length $ L $, and $ H(\cdot) $ represents the entropy function:
\begin{equation}
    H(z) = - \sum_{i} p_i \log p_i, \ \ \text{where } p_i = \frac{\exp(\hat{y}_i)}{\sum_j \exp(\hat{y}_i)}.
\end{equation}
This approach enables the identification of a suboptimal $\mr{L}_{\text{pg}}$ value at test time with a feasible number of forward passes. 

\subsubsection{Normalization Statistics Update} In \acrfull{tta}, the process typically consists of two distinct passes: \textbf{1) Adaptation}, where a subset of parameters $\theta$ is updated based on a target sample $\XX_j^T$ and a test-time objective; and \textbf{2) Inference}, where accuracy is evaluated on the same sample $\XX_j^T$. During the adaptation phase, the model $ f_\theta $ benefits implicitly from BatchNorm statistic updates \cite{ioffe2015batch} in addition to explicit parameter updates.

In contrast, our backpropagation-free method lacks an adaptation pass, and token purging occurs solely during the inference phase. Consequently, our approach does not benefit from BatchNorm statistic updates. To mitigate this limitation, we reset BatchNorm statistics before each forward pass, ensuring that it utilizes the current batch statistics rather than the accumulated and potentially incompatible train-time statistics. For further details, refer to \cite{ioffe2015batch}.

\section{Experiments}
\label{sec:experiments}
This section provides a detailed evaluation of the proposed method across multiple 3D point cloud datasets. To assess its robustness and generalization capabilities, we perform experiments on three benchmark datasets: ModelNet40-C \cite{sun2022benchmarking}, ShapeNet-C \cite{chang2015shapenet}, and ScanObjectNN-C \cite{uy2019revisiting}. These datasets incorporate a range of real-world challenges, including varying degrees of corruption and noise, allowing us to demonstrate the robustness of our approach across diverse and complex conditions. Additionally, we assess the performance of our method in standard adaptation settings, where the model is reset after each batch adaptation.

\subsection{Implementation Details}
For adaptation, we have used the publicly available pre-trained PointMAE \cite{pang2022masked} model form MATE \cite{mirza2023mate} for all the reproduced results to ensure a fair and consistent comparison. The adaptation process was carried out with a batch size of 32 and a single iteration in the standard mode. We have set $\mathcal{L}_{\text{pg}} = \{0, 2, 4, 8, 16, 32\}$ across all the experiments. Since our method does not involve backpropagation, it eliminates the need for an optimizer, making the adaptation process more efficient. To maintain consistency across all configurations, all experiments were conducted on a single NVIDIA A6000 GPU.

\subsection{Datasets}

\mypar{ModelNet40-C.} ModelNet40-C \cite{sun2022benchmarking} serves as a robustness evaluation dataset for point cloud classification, specifically designed to test the ability of architectures to handle distribution shifts encountered in real-world scenarios. It extends the original ModelNet40 test set by introducing 15 corruption types, grouped into three main categories: transformation, noise, and density variations. These corruptions replicate real-world challenges, such as sensor inaccuracies and LiDAR scan noise, offering a comprehensive framework for assessing model resilience under diverse perturbations.

\vspace{2.5 pt}
\mypar{ShapeNet-C.} ShapeNetCore-v2 \cite{chang2015shapenet} is a large-scale dataset widely used for point cloud classification, comprising 51,127 3D shapes spanning 55 categories. The dataset is structured into three subsets: 70\% for training, 10\% for validation, and 20\% for testing. To evaluate model robustness in real-world scenarios, \cite{mirza2023mate} introduced 15 corruption types to the test set, mirroring those found in ModelNet40-C. These corruptions were generated using an open-source implementation provided by \cite{sun2022benchmarking}, resulting in the modified dataset known as ShapeNet-C.

\vspace{2.5 pt}
\mypar{ScanObjectNN-C.} ScanObjectNN \cite{uy2019revisiting} is a real-world point cloud classification dataset comprising 15 object classes, with 2,309 samples allocated for training and 581 for testing. To create ScanObjectNN-C, we follow the methodology outlined in \cite{mirza2023mate}, applying 15 corruption types exclusively to the test set and focusing on object-level perturbations.

\subsection{Main Results}
Since the pre-trained weights of BFTT3D \cite{wang2024backpropagation} were not available, we reproduced their results using their code and original settings, along with the same pre-trained weights employed in our method.

\begin{table*}[t]
    \centering
    \resizebox{\textwidth}{!}{
        \begin{tabular}{l|ccccccccccccccc|c}
            \toprule
            Method & uni & gauss & backg & impul & upsam & rbf & rbf-inv & den-dec & dens-inc & shear & rot & cut & distort & oclsion & lidar & Mean \\
            \midrule
            Source Only & 21.69 & 18.76 & 16.70 & 18.42 & 22.20 & 45.96 & 46.99 & 71.95 & 69.36 & 48.88 & 35.63 & 72.98 & 49.40 & \phz6.71 & \phz8.78 & 36.96 \\
            \midrule
            \multicolumn{17}{c}{\textbf{\acrfull{ttt}}} \\
             MATE & 27.50&	29.40&	14.30	&22.20&	25.60&	40.80&	42.00&	73.70&	63.20&	45.10&	35.30&	73.30&	45.30&	\phz7.10	&\phz9.30&	36.90  \improve{\phz(-0.06)}\\
             SMART-PC & 27.50&	39.10&	19.30	&21.50	&29.80	&44.20&	48.90&	68.30&	60.20	&49.40&	45.40&	70.10&	49.10&	\phz8.40&	12.20&	39.60  \improve{\phz(+2.64)}\\
            \midrule
            \multicolumn{17}{c}{\textbf{\acrfull{tta} Using Backpropagation}} \\
            TTT-Rot & 30.20 & 34.10 & 16.20 & 22.60 & 22.60 & 32.40 & 32.10 & 45.50 & 45.00 & 34.50 & 29.30 & 47.80 & 36.20 & \phz7.10 & \phz8.10 & 29.58 \improve{\phz(-7.38)} \\
            SHOT & 29.50 & 31.60 & 17.60 & 24.80 & 27.20 & 31.00 & 32.40 & 40.70 & 35.00 & 30.20 & 26.60 & 36.60 & 29.30 & \textbf{10.50} & 12.40 & 27.69 \improve{\phz(-9.27)} \\
            T3A\textsuperscript{\,*}& 35.11 & 46.82 & 13.08 & 30.64 & 38.55 & 50.26 & 53.18 & 68.50 & 63.51 & 50.60 & 46.64 & 69.02 & 53.53 & \phz7.23 & 10.16 & 42.46 \improve{\phz(+5.50)} \\
            TENT\textsuperscript{\,*}& 35.28 & 49.74 & 14.46 & 29.60 & 38.04 & 50.43 & 51.81 & 71.60 & 68.85 & 53.70 & 44.23 & 71.43 & 55.77 & \phz7.57 & \phz9.29 & 43.45 \improve{\phz(+6.49)} \\

            \midrule
            \multicolumn{17}{c}{\textbf{\acrfull{tta} Backpropagation Free}} \\
            DUA & 30.50 & 40.10 & 10.20 & 23.60 & 29.90 & 43.70 & 46.10 & 68.30 & 66.30 & 48.50 & 38.90 & 68.70 & 48.40 & \phz8.60 & \phz8.10 & 38.66 \improve{\phz(+1.70)} \\
            BFTT3D\textsuperscript{\,*}& 16.35 & 14.97 & 14.29 & 17.04 & 13.43 & 38.90 & 42.00 & 69.71 & 70.74 & 41.14 & 28.92 & 69.19 & 43.20 & \phz6.54 & \phz8.61 & 33.00 \improve{\phz(-3.96)} \\
            \colrow
            PG-SP (Ours) & \textbf{41.48} & \textbf{55.59} & \textbf{22.89} & \textbf{38.90} & \textbf{42.17} & \textbf{56.80} & \textbf{62.31} & \textbf{76.08} & \textbf{76.42} & \textbf{59.04} & \textbf{49.74} & \textbf{76.76} & \textbf{62.48} & \phz7.57 & \phz8.78 & \textbf{49.13} \improve{(+12.17)} \\
            \colrow
            PG-SF (Ours) & 40.96 & 54.73 & 21.86 & 36.49 & 39.59 & 55.25 & 58.35 & 75.56 & 74.01 & 56.80 & 48.71 & 75.56 & 58.69 & \phz7.06 & \phz\textbf{9.47} & 47.54 \improve{(+10.58)} \\
            \bottomrule
        \end{tabular}
    }
    \caption{Top-1 Classification Accuracy (\%) for all distribution shifts in the ScanObjectNN-C dataset. The mean improvements over Source Only are shown in parentheses. Methods with\textsuperscript{\,*} are reproduced. \gls{ttt} methods' results are from \cite{Bahri2025SMARTPC}.}
    \label{tab:scanobject}
\end{table*}
\begin{table*}[t]
    \centering
    \resizebox{\textwidth}{!}{
        \begin{tabular}{l|ccccccccccccccc|c}
            \toprule
            Method & uni & gauss & backg & impul & upsam & rbf & rbf-inv & den-dec & dens-inc & shear & rot & cut & distort & oclsion & lidar & Mean \\
            \midrule
            Source Only & 66.53 & 59.12 & 7.21 & 31.69 & 74.72 & 67.71 & 69.77 & 62.28 & 75.08 & 74.35 & 38.13 & 58.63 & 70.02 & 38.53 & 18.64 & 54.16 \\
            \midrule
            \multicolumn{17}{c}{\textbf{\acrfull{ttt}}} \\
            MATE & 69.80&	61.80	&18.90&	63.90&	72.50	&64.00&	66.00&	74.00&	80.80&	71.00	&36.70&	69.20	&66.30& 38.40	&29.90&	58.90 \improve{\phz(+3.99)}\\
            SMART-PC & 82.40&	80.10&	12.00	&67.10	&84.5	&76.00&	78.60	&67.30	&72.90	&73.30& 43.90	&72.60&	73.50	&37.40&	24.80&	63.10 \improve{\phz(+8.94)}\\
            \midrule
            \multicolumn{17}{c}{\textbf{\acrfull{tta} Using Backpropagation}} \\
            TTT-Rot & 61.30 & 58.30 & 34.50 & 48.90 & 66.70 & 63.60 & 63.90 & 59.80 & 68.60 & 55.20 & 27.30 & 54.60 & 64.00 & 40.00 & 29.10 & 53.05 \improve{\phz(-1.11)} \\
            SHOT & 29.60 & 28.20 & 9.80 & 25.40 & 32.70 & 30.30 & 30.10 & 30.90 & 31.20 & 32.10 & 22.80 & 27.30 & 29.40 & 20.80 & 18.60 & 26.60 \improve{(-27.56)} \\
            T3A\textsuperscript{\,*} & 64.06 & 61.55 & 20.14 & 50.12 & 64.83 & 63.74 & 65.03 & 66.57 & 72.29 & 68.03 & 50.49 & 64.67 & 62.80 & 43.64 & 40.28 & 57.22 \improve{\phz(+3.06)} \\
            TENT\textsuperscript{\,*} & 73.78 & 70.14 & 18.48 & 59.68 & 76.14 & 70.58 & 72.08 & 73.22 & 79.25 & 73.95 & 54.25 & 70.99 & 70.50 & 46.07 & 43.80 & 63.53 \improve{\phz(+9.37)} \\

            \midrule
            \multicolumn{17}{c}{\textbf{\acrfull{tta} Backpropagation Free}} \\
            DUA & 65.00 & 58.50 & 14.70 & 48.50 & 68.80 & 62.80 & 63.20 & 62.10 & 66.20 & 68.80 & 46.20 & 53.80 & 64.70 & 41.20 & 36.50 & 54.73 \improve{\phz(+0.57)} \\
            BFTT3D\textsuperscript{\,*} & 66.53 & 59.20 & 7.01 & 31.85 & 65.60 & 67.79 & 69.81 & \textbf{77.96} & \textbf{85.25} & 74.35 & 38.17 & 74.80 & 70.02 & 38.61 & 30.51 & 57.16 \improve{\phz(+3.00)} \\
            \colrow
            PG-SP (Ours) & 77.92 & 73.83 & 28.12 & \textbf{69.00} & 80.88 & 72.93 & \textbf{76.09} & 81.40 & 83.59 & 77.43 & 56.97 & \textbf{79.58} & \textbf{74.60} & \textbf{48.99} & \textbf{52.67} & \textbf{68.93} \improve{(+14.77)} \\
            \colrow
            PG-SF (Ours) & \textbf{78.12} & \textbf{74.47} & \textbf{40.88} & 66.69 & \textbf{81.73} & \textbf{73.58} & 75.81 & 79.09 & 82.37 & \textbf{77.23} & \textbf{58.19} & 75.49 & 73.58 & 47.85 & 45.14 & 68.68 \improve{(+14.52)} \\
            \bottomrule
        \end{tabular}
    }
    \caption{Top-1 Classification Accuracy (\%) for all distribution shifts in the ModelNet-C dataset. The mean improvements over Source Only are shown in parentheses. Methods with\textsuperscript{\,*} are reproduced. \gls{ttt} methods' results are from \cite{Bahri2025SMARTPC}.}
    \label{tab:modelnet}
\end{table*}

\vspace{2.5 pt}
\mypar{ScanObjectNN-C.}  
Table \ref{tab:scanobject} highlights the superiority of \gls{pg-sp} and \gls{pg-sf} over existing methods. \gls{pg-sp} achieves the highest mean accuracy of \textbf{49.13\%}, outperforming all backpropagation-free and backpropagation-based techniques by \textbf{+12.17\%} over Source Only. \gls{pg-sf} achieves \textbf{47.54\%}, maintaining strong performance while being fully source-free. Notably, \gls{pg-sp} surpasses DUA by \textbf{+10.47\%} and BFTT3D by \textbf{+16.13\%}, demonstrating its effectiveness in mitigating domain shift at the token level. It ranks first in 13 out of 15 perturbations, excelling against \textit{uniform noise, gaussian noise, and distortion artifacts}. Similarly, \gls{pg-sf} achieves \textit{state-of-the-art results for a source-free method}, leveraging CLS-token-driven prototype selection without requiring precomputed source statistics.

\vspace{2.5 pt}
\mypar{ModelNet40-C.}  
As shown in Table \ref{tab:modelnet}, \gls{pg-sp} and \gls{pg-sf} achieve the best mean classification accuracies of \textbf{68.93\%} and \textbf{68.68\%}, respectively, exceeding Source Only by \textbf{+14.77\%} and \textbf{+14.52\%}. Both methods outperform backpropagation-based techniques such as TENT (63.53\%) and T3A (57.22\%), proving their efficiency without gradient updates. Compared to backpropagation-free approaches, \gls{pg-sp} improves over BFTT3D by \textbf{+11.77\%} and DUA by \textbf{+14.20\%}, reinforcing the strength of our \textbf{token-purging mechanism}. Our method achieves the highest accuracy in 13 out of 15 corruptions, particularly excelling in \textit{background corruption, uniform noise, and LiDAR distortions}, which typically degrade model performance.

\vspace{2.5 pt}
\mypar{ShapeNet-C.}  
Table \ref{tab:shapenet} demonstrates the effectiveness of \gls{pg-sp} and \gls{pg-sf} in the \textbf{\textit{challenging}} ShapeNet-C dataset, where all other methods perform worse than Source Only. \gls{pg-sf} achieves \textbf{64.40\%} and \gls{pg-sp} \textbf{64.17\%}, marking the \textbf{only improvements} over Source Only, by (\textbf{+3.05\%} and \textbf{+2.82\%}). Both surpass TENT (54.98\%) and T3A (54.18\%) while maintaining computational efficiency. Compared to backpropagation-free methods, \gls{pg-sp} and \gls{pg-sf} outperform DUA (60.77\%) and BFTT3D (60.73\%) by over \textbf{+3.4\%}. \gls{pg-sp} achieves the best accuracy in noise-based corruptions, particularly \textit{impulse noise}, while \gls{pg-sf} excels in \textit{background and rotation variations}. Both variants maintain competitive accuracy across density and geometric transformations, reinforcing their robustness to diverse real-world distribution shifts.

\vspace{2.5pt}
\mypar{Mamba Backbone.}
Table \cref{tab:mamba} demonstrates that our \gls{pg-sp} token-purging strategy transfers seamlessly to the state-space \textit{Point-Mamba} \cite{liang2024pointmamba}. On \textbf{ScanObjectNN-C}, \gls{pg-sp} lifts the mean accuracy from \textbf{29.11\%} to \textbf{41.42\%} (\textbf{+12.31\%}) and ranks first in 13 of 15 corruptions, with large jumps for \textit{uniform noise}, \textit{Gaussian noise}, and \textit{background}. The gains persist on \textbf{ModelNet-C}, where \gls{pg-sp} attains \textbf{68.61\%} mean accuracy—\textbf{+10.29\%} above the source model—while surpassing it on every single perturbation, with the biggest margins on \textit{background} (\textbf{+40.5\%}), \textit{impulse noise} (\textbf{+32.4\%}), and the \textit{cut} corruption (\textbf{+46.6\%}). Although \textbf{ShapeNet-C} remains the most challenging benchmark, \gls{pg-sp} maintains an overall score of \textbf{51.66\%} (-0.26\% versus Source Only) and still alleviates severe shifts such as \textit{background} (\textbf{+27.9\%}) and \textit{impulse noise} (\textbf{+17.2\%}).

\begin{table*}
    \centering
    \resizebox{\textwidth}{!}{
        \begin{tabular}{l|ccccccccccccccc|c}
            \toprule
            Method & uni & gauss & backg & impul & upsam & rbf & rbf-inv & den-dec & dens-inc & shear & rot & cut & distort & oclsion & lidar & Mean \\
            \midrule
            Source Only & 77.34 & 71.78 & 8.64 & 54.51 & 77.90 & 75.50 & 76.06 & \textbf{85.25} & 76.42 & \textbf{80.48} & 57.08 & \textbf{85.12} & 76.05 & 10.97 & 7.13 & 61.35 \\
            \midrule
            \multicolumn{17}{c}{\textbf{\acrfull{ttt}}} \\
            MATE-Standard & {{77.80}} & {{74.70}} &              {4.30} & {66.20} & {{78.60}} & {{76.30}} & {{75.30}} &  {86.1} &  {86.60} & {{79.20}} &             {56.10} &             {84.10} & {{76.10}} &             12.30 & {13.10} & 63.10 \improve{\phz(+1.75)} \\
            SMART-PC-Standard & 80.80	&78.90&	 8.90&	60.40&	81.80&	81.10&	81.70&	84.80&	78.40&	80.80	&63.70&	84.90&	79.80&	11.50&	\phz8.80& 64.40  \improve{\phz(+3.05)}\\
            \midrule
            \multicolumn{17}{c}{\textbf{\acrfull{tta} Using Backpropagation}} \\
            TTT-Rot & 74.60 & 72.40 & 23.10 & 59.90 & 74.90 & 73.80 & 75.00 & 81.40 & 82.00 & 69.20 & 49.10 & 79.90 & 72.70 & \textbf{14.00} & 12.00 & 60.93 \improve{\phz(-0.42)} \\
            SHOT & 44.80 & 42.50 & 12.10 & 37.60 & 45.00 & 43.70 & 44.20 & 48.40 & 49.40 & 45.00 & 32.60 & 46.30 & 39.10 & \phz6.20 & \phz5.90 & 36.19 \improve{(-25.16)} \\
            T3A\textsuperscript{\,*}& 66.67 & 64.25 & 13.02 & 59.77 & 67.02 & 65.14 & 64.89 & 70.40 & 66.67 & 65.98 & 54.96 & 69.43 & 63.79 & 10.07 & 10.59 & 54.18 \improve{\phz(-7.17)} \\
            TENT\textsuperscript{\,*}& 68.50 & 65.29 & \phz7.55 & 55.89 & 67.80 & 65.77 & 66.30 & 76.83 & 68.00 & 69.36 & 51.20 & 76.35 & 64.77 & 10.66 & 10.36 & 54.98 \improve{\phz(-6.37)} \\

            \midrule
            \multicolumn{17}{c}{\textbf{\acrfull{tta} Backpropagation Free}} \\
            DUA & 76.10 & 70.10 & 14.30 & 60.90 & 76.20 & 71.60 & 72.90 & 80.00 & \textbf{83.80} & 77.10 & 57.50 & 75.00 & 72.10 & 11.90 & \textbf{12.10} & 60.77 \improve{\phz(-0.58)} \\
            BFTT3D\textsuperscript{\,*}& 72.58 & 65.91 & 10.57 & 60.86 & 67.78 & 73.45 & 74.38 & 84.55 & 83.52 & 78.58 & 57.96 & 83.71 & 75.17 & 11.95 & 10.00 & 60.73 \improve{\phz(-0.62)} \\
            \colrow
            PG-SP (Ours) & 78.63 & \textbf{76.52} & 15.78 & \textbf{72.70} & 78.79 & 76.49 & \textbf{77.18} & 83.28 & 81.18 & 79.05 & 63.36 & 83.07 & \textbf{76.33} & \phz9.85 & 10.34 & 64.17 \improve{\phz(+2.82)} \\
            \colrow
            PG-SF (Ours) & \textbf{79.21} & 75.43 & \textbf{22.46} & 68.38 & \textbf{78.95} & \textbf{76.78} & 77.12 & 83.52 & 80.73 & 79.24 & \textbf{64.91} & 83.55 & 76.08 & 10.15 & \phz9.55 & \textbf{64.40} \improve{\phz(+3.05)} \\
            \bottomrule
        \end{tabular}
    }
    \caption{Top-1 Classification Accuracy (\%) for all distribution shifts in the ShapeNet-C dataset. The mean improvements over Source Only are shown in parentheses. Methods with\textsuperscript{\,*} are reproduced. \gls{ttt} methods' results are from \cite{Bahri2025SMARTPC}.}
    \label{tab:shapenet}
\end{table*}
\begin{table*}
    \centering
    \resizebox{\textwidth}{!}{
        \begin{tabular}{l|ccccccccccccccc|c}
            \toprule
            Method & uni & gauss & backg & impul & upsam & rbf & rbf-inv & den-dec & dens-inc & shear & rot & cut & distort & oclsion & lidar & Mean \\
            \midrule
            \multicolumn{17}{c}{\textbf{ScanObjectNN-C}} \\
            Source Only & 17.73 & 18.59 & 10.67 & 11.19 & 18.93 & 32.87 & 35.11 & 59.90 & 49.05 & 36.32 & 27.71 & 62.48 & 35.80 & \textbf{9.64} & \textbf{10.67} & 29.11 \\
            \colrow
            PG-SP (Ours)& \textbf{32.29} & \textbf{44.79} & \textbf{25.00} & \textbf{33.85} & \textbf{35.07} & \textbf{45.83} & \textbf{46.53} & \textbf{69.62} & \textbf{63.37} & \textbf{50.35} & \textbf{41.15} & \textbf{67.71} & \textbf{50.00} & 6.94 & 8.85 & \textbf{41.42} \improve{\phz(+12.31)}\\
            \midrule
            \multicolumn{17}{c}{\textbf{ModelNet-C}} \\
            Source Only & 71.35 & 64.59 & 5.35  & 39.26 & 78.53 & 67.30 & 69.90 & 79.94 & \textbf{85.62} & 68.44 & 35.78 & 76.66 & 68.44 & 36.35 & 27.35 & 58.32 \\
            \colrow
            PG-SP (Ours)& \textbf{76.79} & \textbf{73.05} & \textbf{45.82} & \textbf{71.71} & \textbf{79.55} & \textbf{72.36} & \textbf{74.23} & \textbf{84.42} & 85.02 & \textbf{69.44} & \textbf{49.31} & \textbf{82.43} & \textbf{71.23} & \textbf{46.83} & \textbf{47.04} & \textbf{68.61} \improve{\phz(+10.29)}\\
            \midrule
            \multicolumn{17}{c}{\textbf{ShapeNet-C}} \\
            Source Only & \textbf{67.20} & \textbf{61.20} & 4.93 & 38.79 & \textbf{67.31} & \textbf{66.55} & \textbf{68.06} & \textbf{80.19} & \textbf{59.26} & \textbf{70.26} & 37.45 & \textbf{80.93} & \textbf{65.96} & 5.96 & 4.71 & \textbf{51.92} \\
            \colrow
            PG-SP (Ours)& 62.89 & 52.26 & \textbf{32.82} & \textbf{56.04} & 64.38 & 62.41 & 64.18 & 70.82 & 57.09 & 62.37 & \textbf{41.25} & 69.54 & 61.38 & \textbf{8.09} & \textbf{9.35} & 51.66 \improve{\phz(-0.26)}\\
            \bottomrule
        \end{tabular}
    }
    \caption{Top-1 Classification Accuracy (\%) on Mamba networks for all distribution shifts in ScanObjectNN-C, ModelNet-C, and ShapeNet-C datasets. The mean improvements over Source Only are shown in parentheses. The Mamba backbone is from \cite{liang2024pointmamba}.}
    \label{tab:mamba}
\end{table*}

\subsection{Ablation Study}
\mypar{Effect of Purge Size $\mathcal{L}_{\text{pg}}$ on the Performance.} 
The purge size \(\mathcal{L}_{\text{pg}}\) is the sole hyperparameter in \gls{pg}, controlling the number of tokens removed before they enter the attention layers. Figure~\ref{fig:purge_size_combined} presents the impact of varying \(\mathcal{L}_{\text{pg}}\), where the black curve represents a dense evaluation across values from 0 to 127, spanning 127 independent runs. Additionally, the red curve depicts the corresponding average entropy of the logits for each run. As the purge size increases ($ 0 < L_{\text{pg}} < 100$), both TOP-1 ACCs steadily improve, and logit entropy decreases, demonstrating that removing highly corrupted tokens enhances model confidence and overall adaptation performance. Excessive purging ($L_{\text{pg}} \gtrsim 100$) causes the removal of clean, informative tokens, resulting in a performance drop along with an increase in entropy.
The green dashed line in Figure~\ref{fig:purge_size_combined} indicates the maximum accuracy attainable within the dense exploration of different purge sizes in the range of \([0, 32]\). However, our entropy-based selection strategy (the orange dashed line) achieves a close to perfect accuracy while operating over a sparse set of only six purge sizes, \(\mathcal{L}_{\text{pg}} = \{0, 2, 4, 8, 16, 32\}\). These results validate the effectiveness of our entropy-based approach in dynamically selecting a near-optimal \(L_{\text{pg}}\), ensuring robust adaptation without requiring exhaustive hyperparameter tuning. This evaluation is conducted on the \textit{Background} corruption from the ScanObjectNN-C dataset.

\vspace{2.5 pt}
\mypar{Effect of BatchNorm Reset.}
 
We have ablated the effect of resetting BatchNorm statistics on our \gls{pg-sp} method and our baseline, BFTT3D.
The results indicate that resetting BatchNorm statistics consistently improves adaptation performance. Specifically, our \gls{pg-sp} model achieves \textbf{49.13\%} accuracy with BatchNorm reset compared to \textbf{39.86\%} without it, showing a substantial performance gain of (+9.27\%). A similar trend is observed for BFTT3D, where BatchNorm reset improves accuracy from \textbf{33.00\%} to \textbf{45.76\%} (+12.76\%). 
Notably, in both cases, our \gls{pg-sp} method outperforms BFTT3D, achieving a higher accuracy with and without BatchNorm reset.

\vspace{2.5 pt}
\mypar{Computational Cost.}
Our \gls{pg-sp} method demonstrates significant computational efficiency compared to BFTT3D in both execution time and memory usage. 
gls{pg-sp} processes a batch in just \textbf{14.07 ms}, whereas BFTT3D requires \textbf{173.94 ms}, achieving a \textbf{12.4×} speedup. For the sake of fairness, we run our Test-Time Hyperparameter Selection is sequential mode and not in parallel.
Additionally, running \gls{pg-sp} consumes \textbf{1300 MiB} of GPU memory in total (backbone and data flow), while BFTT3D requires \textbf{7400 MiB}, resulting in a \textbf{5.5×} lower memory footprint.

For more detailed ablations, please refer to the supplementary material.


\section{Conclusions}
\label{sec:conclusions}
\acrfull{pg} is a backpropagation-free test-time adaptation method that removes tokens most affected by domain shifts, ensuring robust inference for 3D point cloud classification. Our approach outperforms existing TTA methods, achieving higher accuracy (+10.3\%), 12.4× faster inference, and 5.5× lower memory usage. Extensive evaluations demonstrate that token-level purging is a powerful and efficient alternative to traditional fine-tuning, free from backpropagation. Eliminating the need for iterative updates, \gls{pg} suits real-time adaptation under domain shift for real-world applications.


\bibliographystyle{unsrtnat}
\bibliography{main}  

\begin{thebibliography}{43}
\providecommand{\natexlab}[1]{#1}
\providecommand{\url}[1]{\texttt{#1}}
\expandafter\ifx\csname urlstyle\endcsname\relax
  \providecommand{\doi}[1]{doi: #1}\else
  \providecommand{\doi}{doi: \begingroup \urlstyle{rm}\Url}\fi

\bibitem[Qi et~al.(2017{\natexlab{a}})Qi, Su, Mo, and Guibas]{qi2017pointnet}
Charles~R Qi, Hao Su, Kaichun Mo, and Leonidas~J Guibas.
\newblock {PointNet}: {Deep} learning on point sets for {3D} classification and segmentation.
\newblock In \emph{Proceedings of the IEEE conference on computer vision and pattern recognition}, pages 652--660, 2017{\natexlab{a}}.

\bibitem[Qi et~al.(2017{\natexlab{b}})Qi, Yi, Su, and Guibas]{qi2017pointnet++}
Charles~Ruizhongtai Qi, Li~Yi, Hao Su, and Leonidas~J Guibas.
\newblock {PointNet++}: {Deep} hierarchical feature learning on point sets in a metric space.
\newblock \emph{Advances in neural information processing systems}, 30, 2017{\natexlab{b}}.

\bibitem[Wang et~al.(2019)Wang, Sun, Liu, Sarma, Bronstein, and Solomon]{wang2019dynamic}
Yue Wang, Yongbin Sun, Ziwei Liu, Sanjay~E Sarma, Michael~M Bronstein, and Justin~M Solomon.
\newblock Dynamic graph cnn for learning on point clouds.
\newblock \emph{ACM Transactions on Graphics (tog)}, 38\penalty0 (5):\penalty0 1--12, 2019.

\bibitem[Pang et~al.(2022)Pang, Wang, Tay, Liu, Tian, and Yuan]{pang2022masked}
Yatian Pang, Wenxiao Wang, Francis~EH Tay, Wei Liu, Yonghong Tian, and Li~Yuan.
\newblock Masked autoencoders for point cloud self-supervised learning.
\newblock In \emph{European conference on computer vision}, pages 604--621. Springer, 2022.

\bibitem[Zhang et~al.(2022{\natexlab{a}})Zhang, Guo, Gao, Fang, Zhao, Wang, Qiao, and Li]{zhang2022point}
Renrui Zhang, Ziyu Guo, Peng Gao, Rongyao Fang, Bin Zhao, Dong Wang, Yu~Qiao, and Hongsheng Li.
\newblock Point-{M2AE}: {Multi}-scale masked autoencoders for hierarchical point cloud pre-training.
\newblock \emph{Advances in neural information processing systems}, 35:\penalty0 27061--27074, 2022{\natexlab{a}}.

\bibitem[Zhang et~al.(2023)Zhang, Wang, Qiao, Gao, and Li]{zhang2023learning}
Renrui Zhang, Liuhui Wang, Yu~Qiao, Peng Gao, and Hongsheng Li.
\newblock Learning {3D} representations from {2D} pre-trained models via image-to-point masked autoencoders.
\newblock In \emph{Proceedings of the IEEE/CVF Conference on Computer Vision and Pattern Recognition}, pages 21769--21780, 2023.

\bibitem[Bahri et~al.(2024{\natexlab{a}})Bahri, Yazdanpanah, Noori, Cheraghalikhani, Hakim, Osowiechi, Beizaee, Ayed, and Desrosiers]{bahri2024geomask3d}
Ali Bahri, Moslem Yazdanpanah, Mehrdad Noori, Milad Cheraghalikhani, Gustavo Adolfo~Vargas Hakim, David Osowiechi, Farzad Beizaee, Ismail~Ben Ayed, and Christian Desrosiers.
\newblock {GeoMask3D}: {Geometrically} informed mask selection for self-supervised point cloud learning in {3D}.
\newblock \emph{arXiv preprint arXiv:2405.12419}, 2024{\natexlab{a}}.

\bibitem[Gandelsman et~al.(2022)Gandelsman, Sun, Chen, and Efros]{gandelsman2022test}
Yossi Gandelsman, Yu~Sun, Xinlei Chen, and Alexei Efros.
\newblock Test-time training with masked autoencoders.
\newblock \emph{Advances in Neural Information Processing Systems}, 35:\penalty0 29374--29385, 2022.

\bibitem[Gao et~al.(2023)Gao, Zhang, Liu, Darrell, Shelhamer, and Wang]{gao2023back}
Jin Gao, Jialing Zhang, Xihui Liu, Trevor Darrell, Evan Shelhamer, and Dequan Wang.
\newblock Back to the source: Diffusion-driven adaptation to test-time corruption.
\newblock In \emph{Proceedings of the IEEE/CVF Conference on Computer Vision and Pattern Recognition}, pages 11786--11796, 2023.

\bibitem[Iwasawa and Matsuo(2021)]{iwasawa2021test}
Yusuke Iwasawa and Yutaka Matsuo.
\newblock Test-time classifier adjustment module for model-agnostic domain generalization.
\newblock \emph{Advances in Neural Information Processing Systems}, 34:\penalty0 2427--2440, 2021.

\bibitem[Lim et~al.(2023)Lim, Kim, Choo, and Choi]{lim2023ttn}
Hyesu Lim, Byeonggeun Kim, Jaegul Choo, and Sungha Choi.
\newblock {TTN}: {A} domain-shift aware batch normalization in test-time adaptation.
\newblock \emph{arXiv preprint arXiv:2302.05155}, 2023.

\bibitem[Yeo et~al.(2023)Yeo, Kar, Sodagar, and Zamir]{yeo2023rapid}
Teresa Yeo, O{\u{g}}uzhan~Fatih Kar, Zahra Sodagar, and Amir Zamir.
\newblock Rapid network adaptation: Learning to adapt neural networks using test-time feedback.
\newblock In \emph{Proceedings of the IEEE/CVF International Conference on Computer Vision}, pages 4674--4687, 2023.

\bibitem[Schneider et~al.(2020)Schneider, Rusak, Eck, Bringmann, Brendel, and Bethge]{schneider2020improving}
Steffen Schneider, Evgenia Rusak, Luisa Eck, Oliver Bringmann, Wieland Brendel, and Matthias Bethge.
\newblock Improving robustness against common corruptions by covariate shift adaptation.
\newblock \emph{Advances in neural information processing systems}, 33:\penalty0 11539--11551, 2020.

\bibitem[Yazdanpanah et~al.(2022)Yazdanpanah, Rahman, Chaudhary, Desrosiers, Havaei, Belilovsky, and Kahou]{yazdanpanah2022revisiting}
Moslem Yazdanpanah, Aamer~Abdul Rahman, Muawiz Chaudhary, Christian Desrosiers, Mohammad Havaei, Eugene Belilovsky, and Samira~Ebrahimi Kahou.
\newblock Revisiting learnable affines for batch norm in few-shot transfer learning.
\newblock In \emph{Proceedings of the IEEE/CVF conference on computer vision and pattern recognition}, pages 9109--9118, 2022.

\bibitem[Yazdanpanah and Moradi(2022)]{yazdanpanah2022visual}
Moslem Yazdanpanah and Parham Moradi.
\newblock Visual domain bridge: A source-free domain adaptation for cross-domain few-shot learning.
\newblock In \emph{Proceedings of the IEEE/CVF conference on computer vision and pattern recognition}, pages 2868--2877, 2022.

\bibitem[Osowiechi et~al.(2024{\natexlab{a}})Osowiechi, Hakim, Noori, Cheraghalikhani, Bahri, Yazdanpanah, Ben~Ayed, and Desrosiers]{osowiechi2024nc}
David Osowiechi, Gustavo A~Vargas Hakim, Mehrdad Noori, Milad Cheraghalikhani, Ali Bahri, Moslem Yazdanpanah, Ismail Ben~Ayed, and Christian Desrosiers.
\newblock Nc-ttt: A noise constrastive approach for test-time training.
\newblock In \emph{Proceedings of the IEEE/CVF Conference on Computer Vision and Pattern Recognition}, pages 6078--6086, 2024{\natexlab{a}}.

\bibitem[Osowiechi et~al.(2024{\natexlab{b}})Osowiechi, Noori, Vargas~Hakim, Yazdanpanah, Bahri, Cheraghalikhani, Dastani, Beizaee, Ayed, and Desrosiers]{osowiechi2024watt}
David Osowiechi, Mehrdad Noori, Gustavo Vargas~Hakim, Moslem Yazdanpanah, Ali Bahri, Milad Cheraghalikhani, Sahar Dastani, Farzad Beizaee, Ismail Ayed, and Christian Desrosiers.
\newblock Watt: Weight average test time adaptation of clip.
\newblock \emph{Advances in neural information processing systems}, 37:\penalty0 48015--48044, 2024{\natexlab{b}}.

\bibitem[Hakim et~al.(2025)Hakim, Osowiechi, Noori, Cheraghalikhani, Bahri, Yazdanpanah, Ayed, and Desrosiers]{hakim2025clipartt}
Gustavo A~Vargas Hakim, David Osowiechi, Mehrdad Noori, Milad Cheraghalikhani, Ali Bahri, Moslem Yazdanpanah, Ismail~Ben Ayed, and Christian Desrosiers.
\newblock Clipartt: Adaptation of clip to new domains at test time.
\newblock In \emph{2025 IEEE/CVF Winter Conference on Applications of Computer Vision (WACV)}, pages 7092--7101. IEEE, 2025.

\bibitem[Noori et~al.(2025)Noori, Osowiechi, Hakim, Bahri, Yazdanpanah, Dastani, Beizaee, Ayed, and Desrosiers]{noori2025test}
Mehrdad Noori, David Osowiechi, Gustavo Adolfo~Vargas Hakim, Ali Bahri, Moslem Yazdanpanah, Sahar Dastani, Farzad Beizaee, Ismail~Ben Ayed, and Christian Desrosiers.
\newblock Test-time adaptation of vision-language models for open-vocabulary semantic segmentation.
\newblock \emph{arXiv preprint arXiv:2505.21844}, 2025.

\bibitem[Nado et~al.(2020)Nado, Padhy, Sculley, D'Amour, Lakshminarayanan, and Snoek]{nado2020evaluating}
Zachary Nado, Shreyas Padhy, D~Sculley, Alexander D'Amour, Balaji Lakshminarayanan, and Jasper Snoek.
\newblock Evaluating prediction-time batch normalization for robustness under covariate shift.
\newblock \emph{arXiv preprint arXiv:2006.10963}, 2020.

\bibitem[Wang et~al.(2020)Wang, Shelhamer, Liu, Olshausen, and Darrell]{wang2020tent}
Dequan Wang, Evan Shelhamer, Shaoteng Liu, Bruno Olshausen, and Trevor Darrell.
\newblock Tent: Fully test-time adaptation by entropy minimization.
\newblock \emph{arXiv preprint arXiv:2006.10726}, 2020.

\bibitem[Boudiaf et~al.(2022)Boudiaf, Mueller, Ben~Ayed, and Bertinetto]{boudiaf2022parameter}
Malik Boudiaf, Romain Mueller, Ismail Ben~Ayed, and Luca Bertinetto.
\newblock Parameter-free online test-time adaptation.
\newblock In \emph{Proceedings of the IEEE/CVF Conference on Computer Vision and Pattern Recognition}, pages 8344--8353, 2022.

\bibitem[Zhang et~al.(2022{\natexlab{b}})Zhang, Levine, and Finn]{zhang2022memo}
Marvin Zhang, Sergey Levine, and Chelsea Finn.
\newblock Memo: Test time robustness via adaptation and augmentation.
\newblock \emph{Advances in neural information processing systems}, 35:\penalty0 38629--38642, 2022{\natexlab{b}}.

\bibitem[Wang et~al.(2024{\natexlab{a}})Wang, Cheraghian, Hayder, Hong, Ramasinghe, Rahman, Ahmedt-Aristizabal, Li, Petersson, and Harandi]{wang2024backpropagation}
Yanshuo Wang, Ali Cheraghian, Zeeshan Hayder, Jie Hong, Sameera Ramasinghe, Shafin Rahman, David Ahmedt-Aristizabal, Xuesong Li, Lars Petersson, and Mehrtash Harandi.
\newblock Backpropagation-free network for 3d test-time adaptation.
\newblock In \emph{Proceedings of the IEEE/CVF Conference on Computer Vision and Pattern Recognition}, pages 23231--23241, 2024{\natexlab{a}}.

\bibitem[Mirza et~al.(2023)Mirza, Shin, Lin, Schriebl, Sun, Choe, Kozinski, Possegger, Kweon, Yoon, et~al.]{mirza2023mate}
M~Jehanzeb Mirza, Inkyu Shin, Wei Lin, Andreas Schriebl, Kunyang Sun, Jaesung Choe, Mateusz Kozinski, Horst Possegger, In~So Kweon, Kuk-Jin Yoon, et~al.
\newblock Mate: Masked autoencoders are online 3d test-time learners.
\newblock In \emph{Proceedings of the IEEE/CVF International Conference on Computer Vision}, pages 16709--16718, 2023.

\bibitem[Wang et~al.(2024{\natexlab{b}})Wang, Hong, Cheraghian, Rahman, Ahmedt-Aristizabal, Petersson, and Harandi]{wang2024continual}
Yanshuo Wang, Jie Hong, Ali Cheraghian, Shafin Rahman, David Ahmedt-Aristizabal, Lars Petersson, and Mehrtash Harandi.
\newblock Continual test-time domain adaptation via dynamic sample selection.
\newblock In \emph{Proceedings of the IEEE/CVF Winter Conference on Applications of Computer Vision}, pages 1701--1710, 2024{\natexlab{b}}.

\bibitem[Bahri et~al.(2024{\natexlab{b}})Bahri, Yazdanpanah, Noori, Dastani, Cheraghalikhani, Osowiech, Beizaee, Ayed, Desrosiers, et~al.]{bahri2024test}
Ali Bahri, Moslem Yazdanpanah, Mehrdad Noori, Sahar Dastani, Milad Cheraghalikhani, David Osowiech, Farzad Beizaee, Ismail~Ben Ayed, Christian Desrosiers, et~al.
\newblock Test-time adaptation in point clouds: Leveraging sampling variation with weight averaging.
\newblock \emph{arXiv preprint arXiv:2411.01116}, 2024{\natexlab{b}}.

\bibitem[Vaswani et~al.(2017)Vaswani, Shazeer, Parmar, Uszkoreit, Jones, Gomez, Kaiser, and Polosukhin]{vaswani2017attention}
Ashish Vaswani, Noam Shazeer, Niki Parmar, Jakob Uszkoreit, Llion Jones, Aidan~N Gomez, {\L}ukasz Kaiser, and Illia Polosukhin.
\newblock Attention is all you need.
\newblock \emph{Advances in neural information processing systems}, 30, 2017.

\bibitem[Kim et~al.(2022)Kim, Shen, Thorsley, Gholami, Kwon, Hassoun, and Keutzer]{kim2022learned}
Sehoon Kim, Sheng Shen, David Thorsley, Amir Gholami, Woosuk Kwon, Joseph Hassoun, and Kurt Keutzer.
\newblock Learned token pruning for transformers.
\newblock In \emph{Proceedings of the 28th ACM SIGKDD Conference on Knowledge Discovery and Data Mining}, pages 784--794, 2022.

\bibitem[Gu and Dao(2023)]{gu2023mamba}
Albert Gu and Tri Dao.
\newblock Mamba: Linear-time sequence modeling with selective state spaces.
\newblock \emph{arXiv preprint arXiv:2312.00752}, 2023.

\bibitem[Bahri et~al.(2025{\natexlab{a}})Bahri, Yazdanpanah, Dastani, Noori, Hakim, Osowiechi, Beizaee, Ayed, and Desrosiers]{Bahri2025SMARTPC}
Ali Bahri, Moslem Yazdanpanah, Sahar Dastani, Mehrdad Noori, Gustavo Adolfo~Vargas Hakim, David Osowiechi, Farzad Beizaee, Ismail~Ben Ayed, and Christian Desrosiers.
\newblock {SMART\textendash PC}: Skeletal model adaptation for robust test\textendash time training in point clouds.
\newblock \emph{Proceedings of the 42\textsuperscript{nd} International Conference on Machine Learning (ICML~2025)}, 2025{\natexlab{a}}.

\bibitem[Shin et~al.(2022)Shin, Tsai, Zhuang, Schulter, Liu, Garg, Kweon, and Yoon]{shin2022mm}
Inkyu Shin, Yi-Hsuan Tsai, Bingbing Zhuang, Samuel Schulter, Buyu Liu, Sparsh Garg, In~So Kweon, and Kuk-Jin Yoon.
\newblock Mm-tta: multi-modal test-time adaptation for 3d semantic segmentation.
\newblock In \emph{Proceedings of the IEEE/CVF Conference on Computer Vision and Pattern Recognition}, pages 16928--16937, 2022.

\bibitem[Tang et~al.(2023)Tang, Zhang, Liu, Liu, and Liu]{tang2023dynamic}
Quan Tang, Bowen Zhang, Jiajun Liu, Fagui Liu, and Yifan Liu.
\newblock Dynamic token pruning in plain vision transformers for semantic segmentation.
\newblock In \emph{Proceedings of the IEEE/CVF International Conference on Computer Vision}, pages 777--786, 2023.

\bibitem[Wu et~al.(2023)Wu, Zeng, Wang, and Chen]{wu2023ppt}
Xinjian Wu, Fanhu Zeng, Xiudong Wang, and Xinghao Chen.
\newblock Ppt: Token pruning and pooling for efficient vision transformers.
\newblock \emph{arXiv preprint arXiv:2310.01812}, 2023.

\bibitem[Liu et~al.(2024)Liu, Tian, Zhao, Yu, Xie, Wang, Ye, Jiao, and Liu]{liu2024vmamba}
Yue Liu, Yunjie Tian, Yuzhong Zhao, Hongtian Yu, Lingxi Xie, Yaowei Wang, Qixiang Ye, Jianbin Jiao, and Yunfan Liu.
\newblock Vmamba: Visual state space model.
\newblock \emph{Advances in neural information processing systems}, 37:\penalty0 103031--103063, 2024.

\bibitem[Liang et~al.(2024)Liang, Zhou, Xu, Zhu, Zou, Ye, Tan, and Bai]{liang2024pointmamba}
Dingkang Liang, Xin Zhou, Wei Xu, Xingkui Zhu, Zhikang Zou, Xiaoqing Ye, Xiao Tan, and Xiang Bai.
\newblock Pointmamba: A simple state space model for point cloud analysis.
\newblock \emph{Advances in neural information processing systems}, 37:\penalty0 32653--32677, 2024.

\bibitem[Dastani et~al.(2025)Dastani, Bahri, Yazdanpanah, Noori, Osowiechi, Hakim, Beizaee, Cheraghalikhani, Mondal, Lombaert, et~al.]{dastani2025spectral}
Sahar Dastani, Ali Bahri, Moslem Yazdanpanah, Mehrdad Noori, David Osowiechi, Gustavo Adolfo~Vargas Hakim, Farzad Beizaee, Milad Cheraghalikhani, Arnab~Kumar Mondal, Herve Lombaert, et~al.
\newblock Spectral state space model for rotation-invariant visual representation learning.
\newblock In \emph{Proceedings of the Computer Vision and Pattern Recognition Conference}, pages 23881--23890, 2025.

\bibitem[Bahri et~al.(2025{\natexlab{b}})Bahri, Yazdanpanah, Noori, Dastani, Cheraghalikhani, Hakim, Osowiechi, Beizaee, Ben~Ayed, and Desrosiers]{bahri2025spectral}
Ali Bahri, Moslem Yazdanpanah, Mehrdad Noori, Sahar Dastani, Milad Cheraghalikhani, Gustavo Adolfo~Vargas Hakim, David Osowiechi, Farzad Beizaee, Ismail Ben~Ayed, and Christian Desrosiers.
\newblock Spectral informed mamba for robust point cloud processing.
\newblock In \emph{Proceedings of the Computer Vision and Pattern Recognition Conference}, pages 11799--11809, 2025{\natexlab{b}}.

\bibitem[Zou et~al.(2024)Zou, Yi, Li, and Li]{zou2024closer2}
Yixiong Zou, Shuai Yi, Yuhua Li, and Ruixuan Li.
\newblock A closer look at the {CLS} token for cross-domain few-shot learning.
\newblock \emph{NeurIPS}, 37:\penalty0 85523--85545, 2024.

\bibitem[Ioffe and Szegedy(2015)]{ioffe2015batch}
Sergey Ioffe and Christian Szegedy.
\newblock Batch normalization: Accelerating deep network training by reducing internal covariate shift.
\newblock In \emph{International conference on machine learning}, pages 448--456. pmlr, 2015.

\bibitem[Sun et~al.(2022)Sun, Zhang, Kailkhura, Yu, Xiao, and Mao]{sun2022benchmarking}
Jiachen Sun, Qingzhao Zhang, Bhavya Kailkhura, Zhiding Yu, Chaowei Xiao, and Z~Morley Mao.
\newblock Benchmarking robustness of 3d point cloud recognition against common corruptions.
\newblock \emph{arXiv preprint arXiv:2201.12296}, 2022.

\bibitem[Chang et~al.(2015)Chang, Funkhouser, Guibas, Hanrahan, Huang, Li, Savarese, Savva, Song, Su, et~al.]{chang2015shapenet}
Angel~X Chang, Thomas Funkhouser, Leonidas Guibas, Pat Hanrahan, Qixing Huang, Zimo Li, Silvio Savarese, Manolis Savva, Shuran Song, Hao Su, et~al.
\newblock Shapenet: An information-rich 3d model repository.
\newblock \emph{arXiv preprint arXiv:1512.03012}, 2015.

\bibitem[Uy et~al.(2019)Uy, Pham, Hua, Nguyen, and Yeung]{uy2019revisiting}
Mikaela~Angelina Uy, Quang-Hieu Pham, Binh-Son Hua, Thanh Nguyen, and Sai-Kit Yeung.
\newblock Revisiting point cloud classification: A new benchmark dataset and classification model on real-world data.
\newblock In \emph{Proceedings of the IEEE/CVF international conference on computer vision}, pages 1588--1597, 2019.

\end{thebibliography}






\newpage
\section{Supplementary Material}

\subsection{Effect of Batch Size.}
\begin{figure}[b]
    \centering
    \includegraphics[width=.85\linewidth]{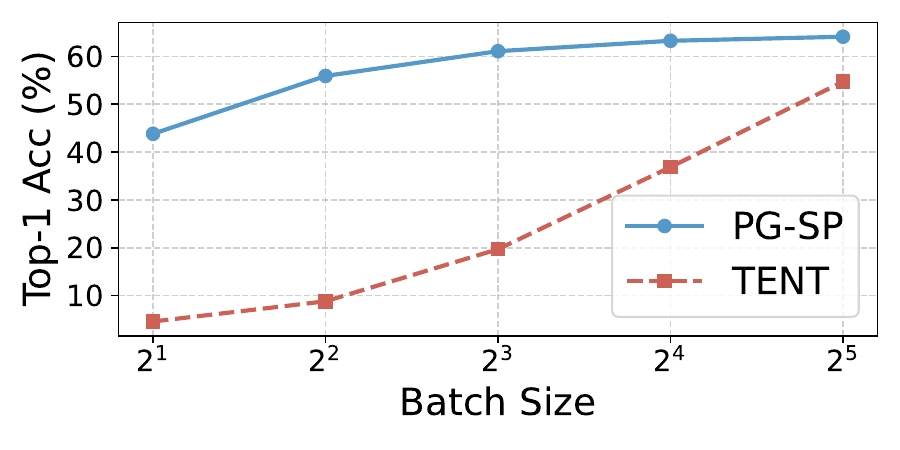}
    \caption{Effect of Batch size on ACC performance of our \gls{pg-sp} and TENT, on ShapeNet-C dataset.}
    \label{fig:batch_size}
\end{figure}

Figure~\ref{fig:batch_size} shows that increasing batch size improves accuracy for both methods, but PG-SP consistently outperforms TENT across all batch sizes. Unlike TENT, which struggles with small batches, PG-SP maintains high accuracy even in low-resource settings, demonstrating its robustness and adaptability.
This highlights a key advantage of PG-SP: it performs well without relying on large batch sizes, making it more practical for real-world deployment with memory or computational constraints. This evaluation is performed on the ShapeNet-C dataset. 

\subsection{Real-world ScanObjectNN variants.}
Table~\ref{tab:scanobjectnn_real} confirms that our \textbf{PG-SP} strategy improves performance even under genuine domain shifts that are free of synthetic corruptions. Starting from the OBJ-ONLY model used in Table 1, we deploy PG-SP on two challenging variants: \textit{OBJ-BG}, which introduces cluttered backgrounds, and \textit{PB-T50-RS}, which combines point-based sampling and rotation. PG-SP raises accuracy from \textbf{74.18\%} to \textbf{75.56\%} on OBJ-BG (\textbf{+1.38\%}), and from textbf{56.77\%} to \textbf{61.07\%} on PB-T50-RS (\textbf{+4.30\%}). These gains demonstrate that our token-purging mechanism benefits real-world scenarios beyond the hand-crafted noise studied in ScanObjectNN-C.

\subsection{Dynamics of Purging and Entropy and its effect on the final ACC}
\begin{figure}[t]
    \centering
    \begin{subfigure}[t]{0.48\textwidth}
        \centering
        \includegraphics[width=\linewidth]{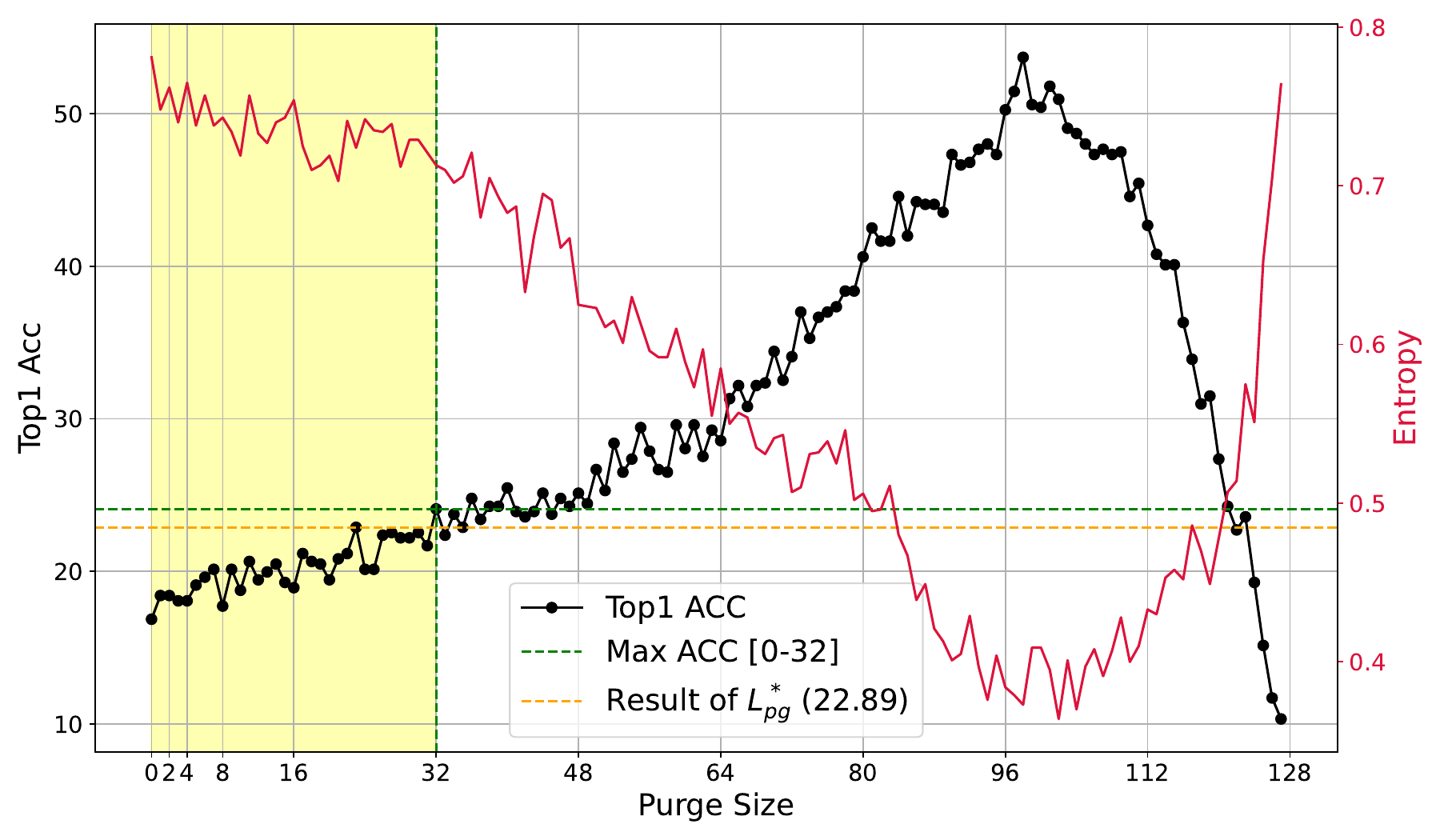}
        \label{fig:purge_size_background}
    \end{subfigure}
    \hfill
    \begin{subfigure}[t]{0.48\textwidth}
        \centering
        \includegraphics[width=\linewidth]{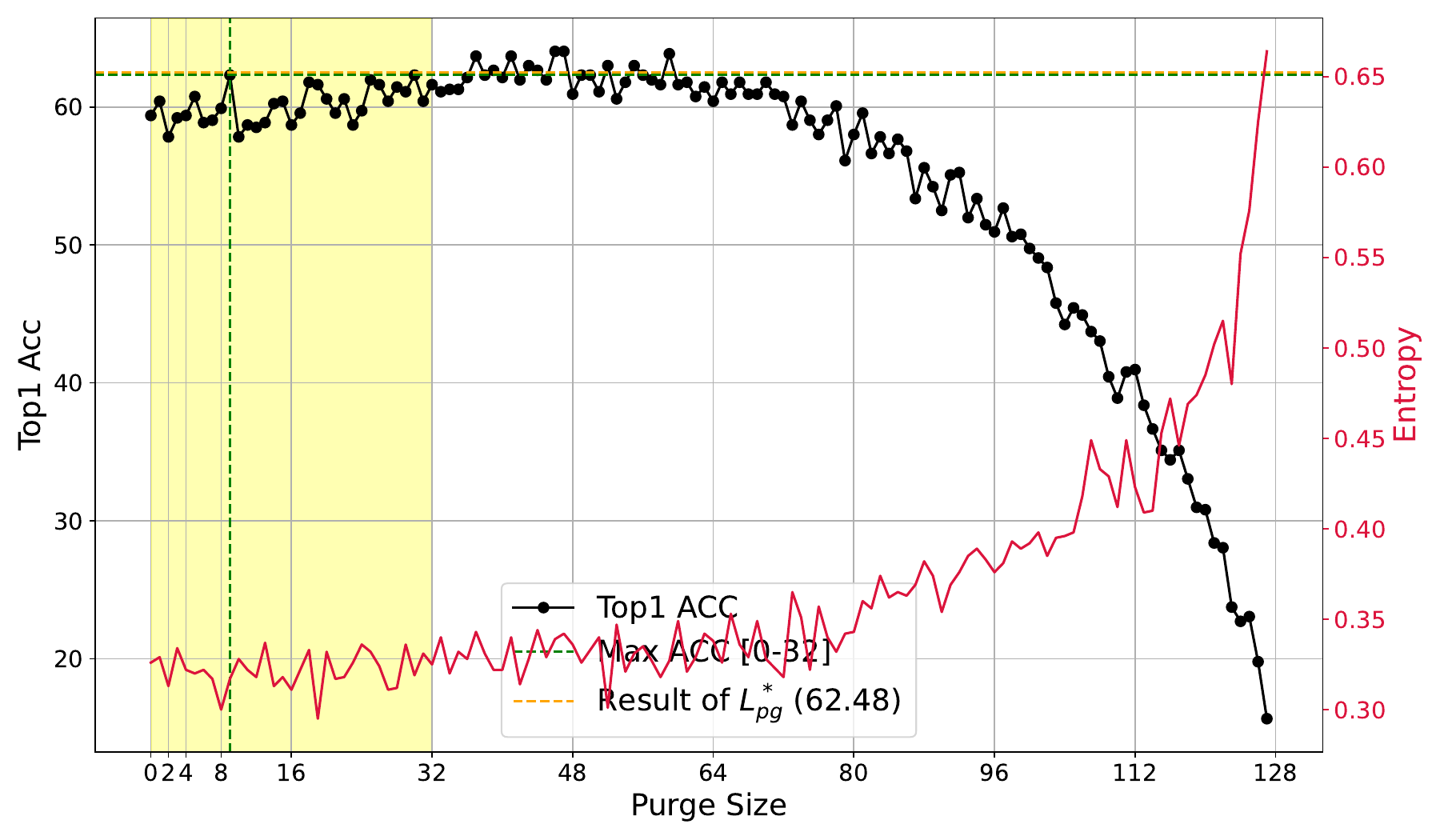}
        \label{fig:purge_size_distortion}
    \end{subfigure}
    \caption{Effect of purge size on ACC under \textbf{Background} (left) and \textbf{Distortion} (right) corruptions from the ScanObjectNN-C dataset.}
    \label{fig:purge_size_combined}
\end{figure}

\vspace{2.5 pt}
\mypar{Effect of purge size $\mathcal{L}_{\text{pg}}$ on ACC under \textit{Distortion} corruption.}
Figure~\ref{fig:purge_size_combined} presents Top-1 accuracy (black, left axis) and logit entropy (red, right axis) for 128 independent runs spanning $\mathcal{L}_{\text{pg}}\!\in\![0,127]$.  
Accuracy starts high at $\mathcal{L}_{\text{pg}}=0$ (59 \%), climbs slightly to a plateau of 63 \% around $\mathcal{L}_{\text{pg}}\!\approx\!48$, then drops sharply once purging exceeds 80, while entropy steadily rises and spikes when informative tokens are removed.  
The green dashed line indicates the maximum accuracy achieved by an exhaustive search restricted to $\mathcal{L}_{\text{pg}}\!\in\![0,32]$.  
The yellow band marks the six purge sizes $\{0,2,4,8,16,32\}$ examined by our entropy-guided schedule; from this sparse set, it selects $\mathcal{L}_{\text{pg}}^{\!*}$ and attains 62.48 \% accuracy (orange dashed line), essentially matching the global optimum without exhaustive tuning, thereby validating entropy as a practical proxy for near-optimal hyper-parameter selection.

\begin{figure}[t]
    \centering
    \begin{subfigure}[t]{0.48\textwidth}
        \centering
        \includegraphics[width=\linewidth]{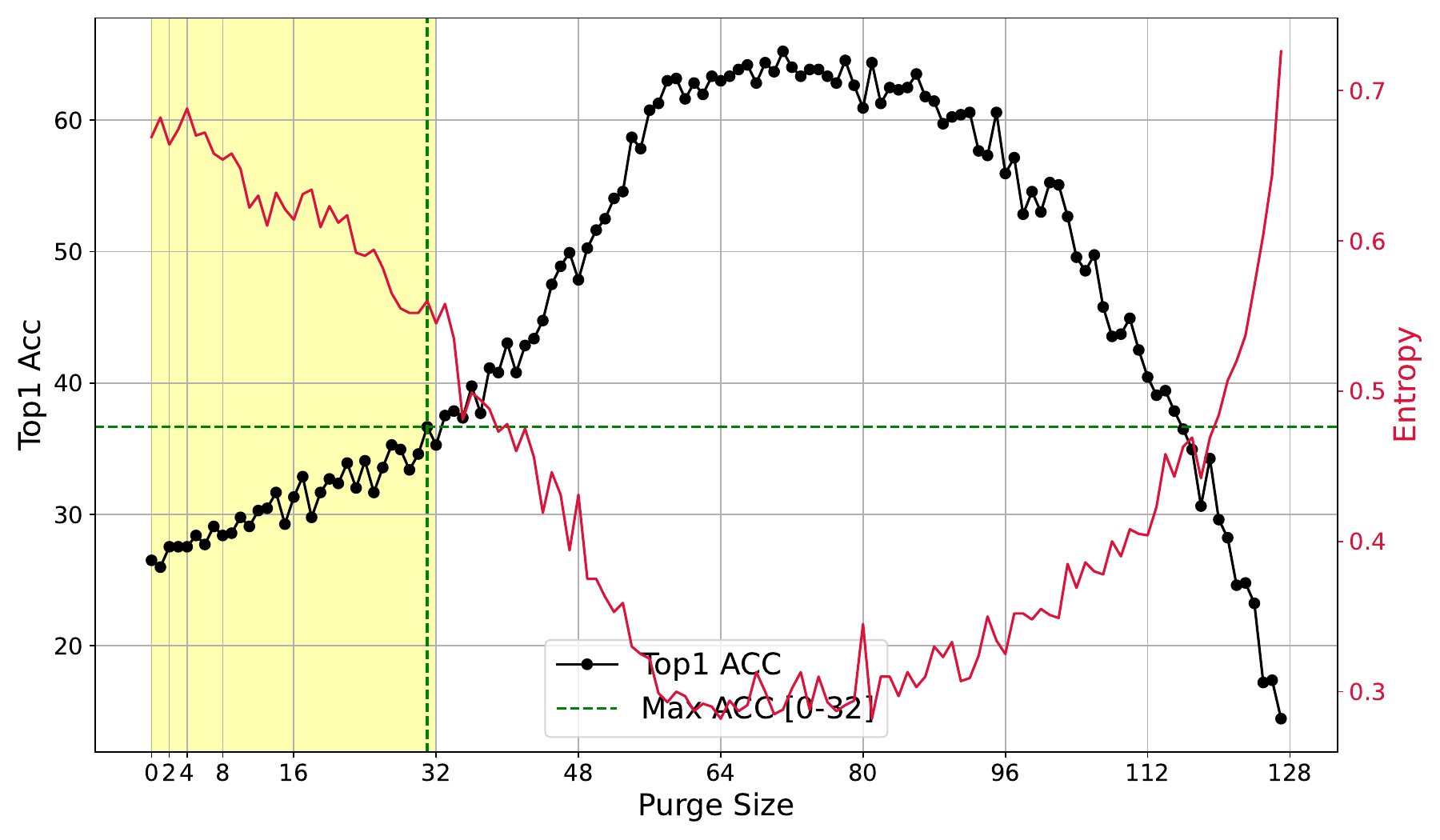}
        \caption{Level 3}
        \label{fig:purge_size_background_3}
    \end{subfigure}
    \hfill
    \begin{subfigure}[t]{0.48\textwidth}
        \centering
        \includegraphics[width=\linewidth]{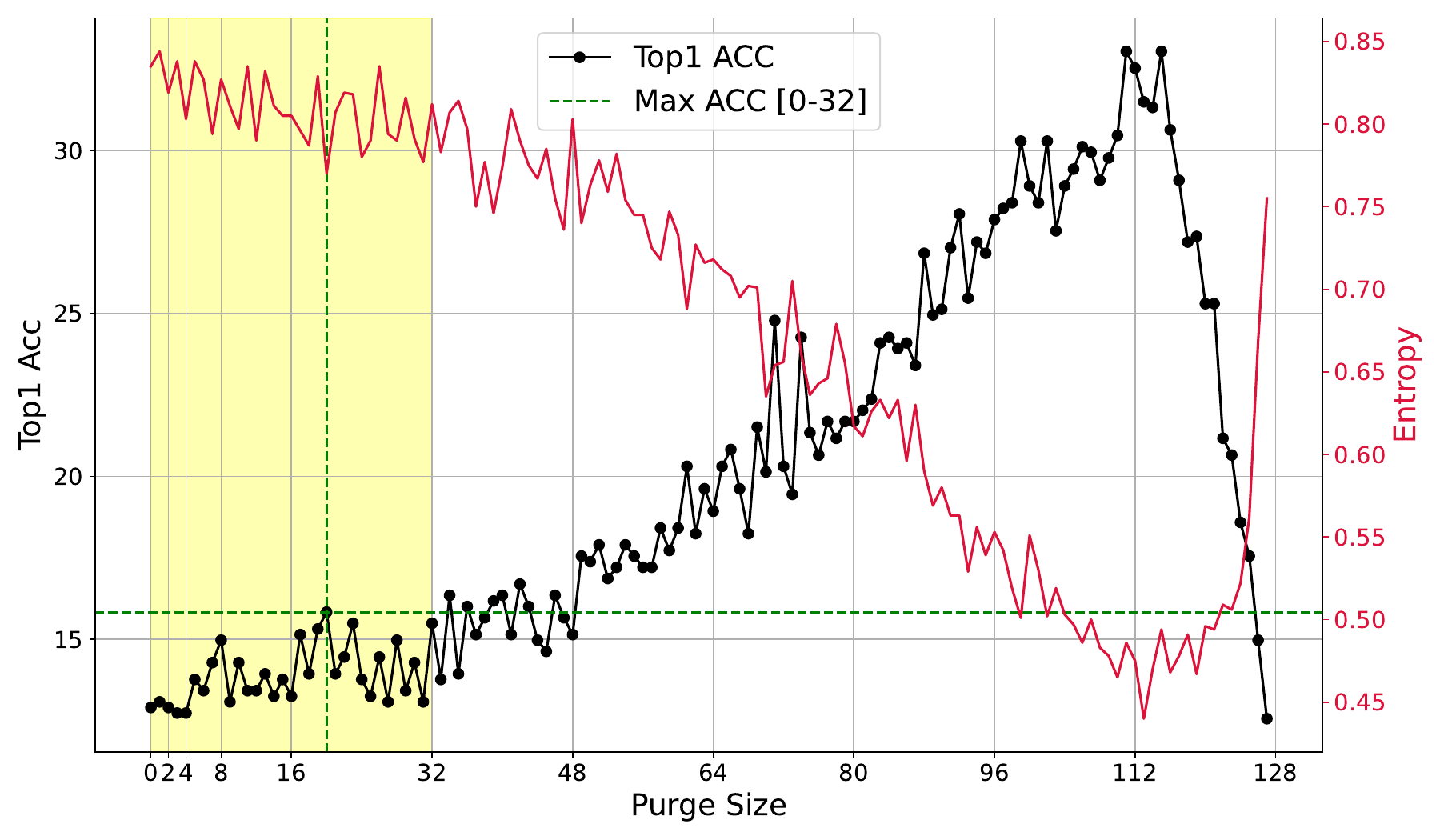}
        \caption{Level 7}
        \label{fig:purge_size_background_7}
    \end{subfigure}
    \caption{Effect of purge size on ACC of \textbf{Background} corruption at different severity levels (3 and 7) from the ScanObjectNN-C dataset.}
    \label{fig:purge_size_background_levels}
\end{figure}

\vspace{2.5 pt}
\mypar{Dynamics of purging: interplay between accuracy and entropy (no $\mathcal{L}_{\text{pg}}$ selection).}
Figure~\ref{fig:purge_size_background_3} traces Top-1 accuracy (black, left axis) and logit entropy (red, right axis) for \textit{Background} corruption level~3 as the purge size $\mathcal{L}_{\text{pg}}$ sweeps the full range $[0,127]$; unlike earlier plots, the result of our entropy-guided schedule is \emph{not} shown.  
Accuracy and entropy evolve inversely: starting at $\mathcal{L}_{\text{pg}}=0$, accuracy is low (27\,\%) while entropy is high (0.60).  Incrementally removing the most corrupted tokens (\,$0<\mathcal{L}_{\text{pg}}\!\lesssim\!64$\,) steadily boosts accuracy to a plateau of 64 \% and drives entropy down to 0.25, indicating increased model confidence.  Beyond $\mathcal{L}_{\text{pg}}\!\gtrsim\!90$, further purging excises informative tokens; entropy rebounds and accuracy collapses, exposing the regime where pruning becomes detrimental.  
The green dashed line marks the best score achievable by an exhaustive yet \emph{narrow} search over $\mathcal{L}_{\text{pg}}\!\in\![0,32]$, yielding only 38 \%—well below the true optimum around $\mathcal{L}_{\text{pg}}\!\approx\!64$.  These dynamics underline the need for a principled, entropy-aware selection strategy: without it, practitioners risk settling on sub-optimal hyper-parameters or over-purging, both of which degrade performance.

\vspace{2.5 pt}
\mypar{Purging dynamics under \textit{Background} corruption, severity 7 (no $\mathcal{L}_{\text{pg}}$ selection shown).}%
Figure~\ref{fig:purge_size_background_7} plots Top-1 accuracy (black, left axis) and logit entropy (red, right axis) across 128 purge sizes $\mathcal{L}_{\text{pg}}\!\in\![0,127]$.  
At $\mathcal{L}_{\text{pg}}=0$, heavy corruption yields low accuracy (13 \%) and high entropy (0.83).  Removing a handful of tokens ($0\!<\!\mathcal{L}_{\text{pg}}\!\le\!32$, yellow band) barely helps—the best accuracy reachable in this restricted range (green dashed line) is only 16 \%.  As purging continues beyond $\mathcal{L}_{\text{pg}}\!\approx\!40$, accuracy climbs almost linearly while entropy declines, peaking near $\mathcal{L}_{\text{pg}}\!\approx\!112$ with 33 \% accuracy and 0.45 entropy.  Past this point the trends reverse: over-purging ($\mathcal{L}_{\text{pg}}\!>\!120$) discards informative tokens, causing accuracy to collapse and entropy to rebound.  These dynamics reveal that severe corruption demands far more aggressive purging than the conventional search window $[0,32]$ can capture; without an adaptive, entropy-aware mechanism, practitioners risk selecting sub-optimal hyper-parameters that leave most potential performance untapped.

\begin{figure}[t]
    \centering
    \begin{subfigure}[t]{0.48\textwidth}
        \centering
        \includegraphics[width=\linewidth]{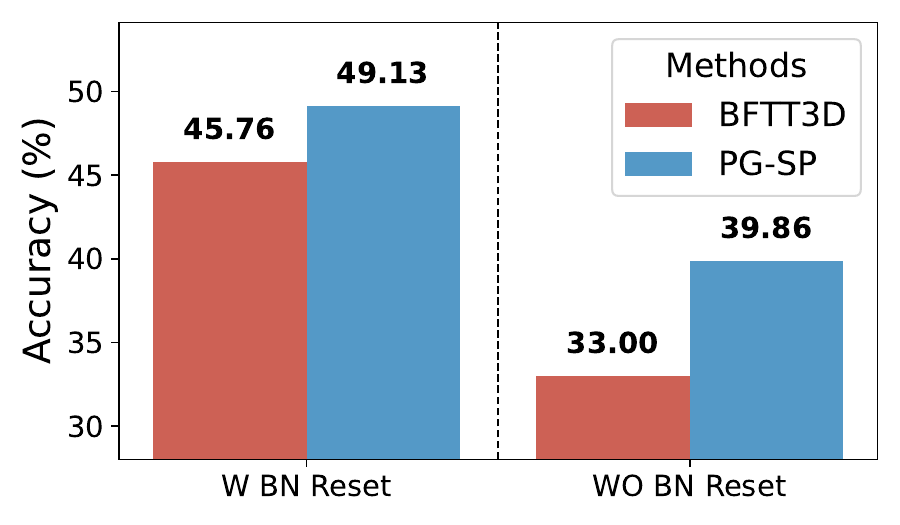}
        \caption{Effect of BatchNorm reset on \gls{pg-sp} and BFTT3D.}
        \label{fig:bn_reset}
    \end{subfigure}
    \hfill
    \begin{subfigure}[t]{0.48\textwidth}
        \centering
        \includegraphics[width=\linewidth]{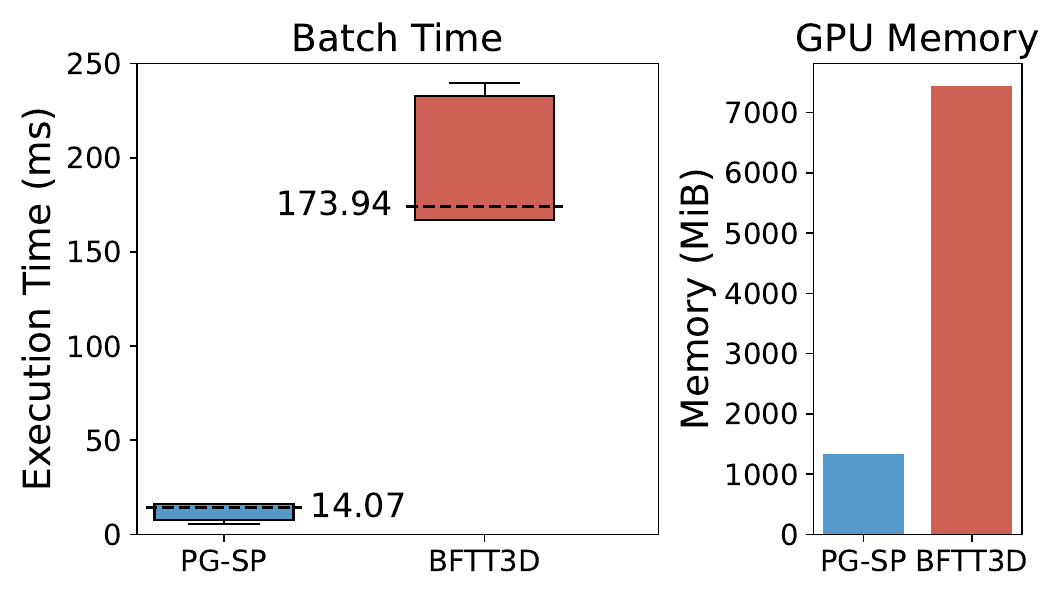}
        \caption{GPU time and memory comparison.}
        \label{fig:computations}
    \end{subfigure}
    \caption{Comparison of our \gls{pg-sp} method and the baseline BFTT3D on the ScanObjectNN-C and ModelNet-C datasets. 
    (a) Effect of BatchNorm reset on accuracy. 
    (b) GPU time and memory usage with batch size 32.}
    \label{fig:bn_reset_time_memory}
\end{figure}

\begin{table}[t]
    \centering
    \resizebox{0.6\linewidth}{!}{
        \begin{tabular}{l|cc|c}
            \toprule
            Variant & Source Only & PG-SP (Ours) & $\Delta$ \\
            \midrule
            OBJ-BG     & 74.18 & \textbf{75.56} & +1.38 \\
            PB-T50-RS  & 56.77 & \textbf{61.07} & +4.30 \\
            \bottomrule
        \end{tabular}
    }
    \caption{Top-1 classification accuracy (\%) on real-world ScanObjectNN variants without synthetic corruptions.}
    \label{tab:scanobjectnn_real}
\end{table}

\end{document}